\documentclass{article}

\PassOptionsToPackage{numbers, compress}{natbib}

\usepackage[final]{neurips_data_2024}
\usepackage{amsmath}
\usepackage[pdftex]{graphicx}
\usepackage{wrapfig}
\usepackage{lipsum}  
\usepackage{xcolor} 
\definecolor{mydarkgreen}{RGB}{0,200,0}
\definecolor{mydarkred}{RGB}{200,0,0}
\usepackage{bbding}
\usepackage{soul}
\usepackage{siunitx}
\usepackage{setspace}
\usepackage{array} 
\usepackage{float}
\usepackage{makecell}
\usepackage{subcaption}
\usepackage{titletoc}      

\usepackage{adjustbox}
\usepackage{colortbl}
\usepackage{comment}
\usepackage{tabularx} 
\usepackage{booktabs} 
\usepackage{multirow} 

\usepackage[utf8]{inputenc} 
\usepackage[T1]{fontenc}    
\usepackage{hyperref}       
\usepackage{url}            
\usepackage{booktabs}       
\usepackage{amsfonts}       
\usepackage{nicefrac}       
\usepackage{microtype}      
\usepackage{xcolor}         

\title{DrivAerNet++: A Large-Scale Multimodal Car Dataset with Computational Fluid Dynamics Simulations and Deep Learning Benchmarks}

\author{%
  Mohamed Elrefaie\thanks{Corresponding author: mohamed.elrefaie@mit.edu} \\
  Department of Mechanical Engineering\\
  Massachusetts Institute of Technology\\
  Cambridge, MA 02139 USA \\
  \And
    Florin Morar \\
  Morphing and Optimization Solutions \\
  BETA CAE SYSTEMS USA, Inc\\
  Farmington Hills, MI 48334 USA\\  \And
    Angela Dai \\
    Department of Computer Science \\
    Technical University of Munich\\
    Garching, 85748 Germany\\
   \And
  Faez Ahmed \\
  Department of Mechanical Engineering\\
  Massachusetts Institute of Technology\\
  Cambridge, MA 02139 USA \\
}

\begin{document}

\maketitle

\begin{abstract}
We present DrivAerNet++, the largest and most comprehensive multimodal dataset for aerodynamic car design. DrivAerNet++ comprises 8,000 diverse car designs modeled with high-fidelity computational fluid dynamics (CFD) simulations. The dataset includes diverse car configurations such as fastback, notchback, and estateback, with different underbody and wheel designs to represent both internal combustion engines and electric vehicles. Each entry in the dataset features detailed 3D meshes, parametric models, aerodynamic coefficients, and extensive flow and surface field data, along with segmented parts for car classification and point cloud data. This dataset supports a wide array of machine learning applications including data-driven design optimization, generative modeling, surrogate model training, CFD simulation acceleration, and geometric classification. With more than 39 TB of publicly available engineering data, DrivAerNet++ fills a significant gap in available resources, providing high-quality, diverse data to enhance model training, promote generalization, and accelerate automotive design processes. 
Along with rigorous dataset validation, we also provide ML benchmarking results on the task of aerodynamic drag prediction, showcasing the breadth of applications supported by our dataset. This dataset is set to significantly impact automotive design and broader engineering disciplines by fostering innovation and improving the fidelity of aerodynamic evaluations. 
Dataset and code available at: \url{https://github.com/Mohamedelrefaie/DrivAerNet}
\end{abstract}
\section{Introduction}

Car design is a complex and iterative process requiring close collaboration between designers, who focus on aesthetics, and engineers, who ensure the design meets performance constraints. One of the key challenges is achieving a balance between aesthetic appeal and aerodynamic efficiency, which directly impacts fuel consumption. With stricter fuel consumption regulations for internal combustion engine (ICE) cars and increased range requirements for battery-powered electric vehicles (BEVs)~\cite{mock2021pathways, brand2020road, martins2023assessing}, ensuring efficient car aerodynamics has become crucial. As a result, there is significant interest in developing machine learning methods for modeling car aerodynamics.

\begin{figure}[t]
    \centering
    \begin{subfigure}[b]{0.49\textwidth}
        \centering
        \includegraphics[height=6.1cm, width=\textwidth]{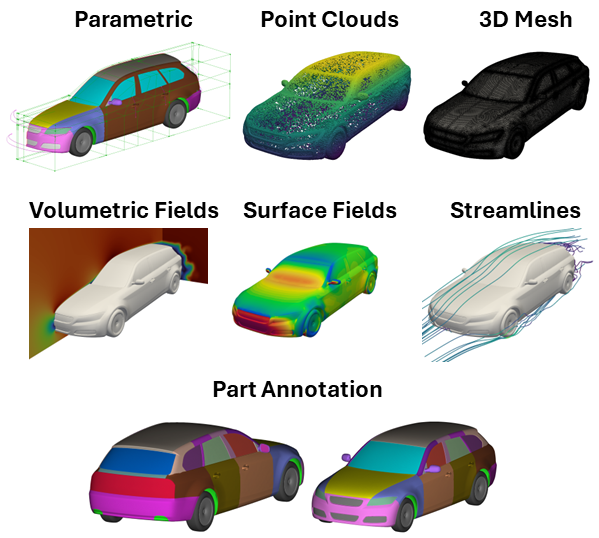}
        \caption{Data modalities of DrivAerNet++. Top row: different data representations; middle row: CFD simulation results; bottom row: annotated car components.\\}
        \label{fig:sub1}
    \end{subfigure}
    \hfill
    \begin{subfigure}[b]{0.49\textwidth}
        \centering
        \includegraphics[height=6cm, width=\textwidth]{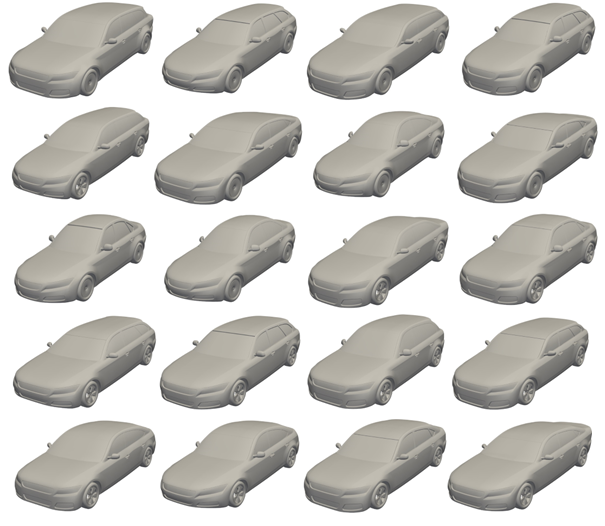}
        \caption{Selected samples from DrivAerNet++ showing diversity in shape with different car designs (fastback, estateback, and notchback), wheels configurations, and underbody configurations.}
        \label{fig:sub2}
    \end{subfigure}
    \caption{Data modalities and shape variations in the DrivAerNet++ dataset.}
    \label{fig:DrivAerNet_intro}
    \vspace{-15pt}
\end{figure}

Data-driven approaches can significantly shorten the process needed before obtaining performance estimates, which typically involves generating the 3D mesh, ensuring watertightness and simulation readiness, performing CFD meshing, defining the solver and boundary conditions, running the CFD, and postprocessing the results. By streamlining these steps, data-driven methods enhance efficiency and expedite the design process. This enables designers to explore various ideas with real-time, accurate performance estimates, ultimately enhancing outcomes with greater design freedom.
Recent advances in geometric deep learning methods~\cite{Rios2021_Point2FFD, Rios2019, remelli2020meshsdf, abbas2022geometrical, kashefi2022physics, trinh20243d, song2023surrogate, baque2018geodesic} have demonstrated their ability to estimate performance values from CFD rapidly, facilitating interactive design modifications. However, these methods are often restricted to simple problems due to a lack of public datasets, which limits their broader applicability.

Existing datasets frequently focus on simpler 2D cases~\cite{bonnet2022airfrans, thuerey2020deep, elrefaie2024surrogate, ayman2023deep, Usama2021, kashefi2022physics, Gunpinar2019} or simplified 3D models~\cite{baque2018geodesic, remelli2020meshsdf, li2023geometryinformed, Umetani2018, Rios2021, song2023surrogate, trinh20243d}, often excluding critical components like wheels, mirrors, and underbodies. As highlighted by~\cite{heft2012experimental}, including these elements significantly affects aerodynamic performance, resulting in a notable increase in drag. This was evidenced by increases in the drag value of approximately 142\% in CFD simulations and 120\% in wind tunnel experiments. This underscores the crucial role of comprehensive 3D modeling in achieving accurate aerodynamic assessments. In addition, around 25$\%$ of a passenger car's aerodynamic drag is attributed directly or indirectly to its wheels~\cite{brandt2019effects}. Furthermore, many large-scale datasets~\cite{song2023surrogate, Umetani2018} lack experimental validation of the baseline simulation with physical wind tunnel tests and validation of the individual convergence of each simulation.

There is a notable lack of public, large-scale, and multimodal car datasets, which hinders progress in data-driven design.  This stands in contrast to other fields, where standardized datasets such as ImageNet~\cite{deng2009imagenet}, ObjectNet3D~\cite{xiang2016objectnet3d}, ModelNet~\cite{wu20153d}, and ScanNet~\cite{dai2017scannet} have driven significant advancements.

We address these challenges with the DrivAerNet++\footnote{\url{https://github.com/Mohamedelrefaie/DrivAerNet}} dataset (see Figure~\ref{fig:DrivAerNet_intro}). 
The DrivAerNet++ dataset represents a significant advancement over its predecessor, the original DrivAerNet dataset~\cite{elrefaie2024drivaernet}
with the integration of 4,000 diverse car shapes to its collection. 
This enhancement doubles the dataset’s volume to a total of 8,000 industry-standard car designs and notably elevates the simulation fidelity with a more complex cell structure (24M cells, as opposed to 8-16M in the original dataset). Furthermore, DrivAerNet++ expands its utility by incorporating detailed 3D flow field data, parametric data, aerodynamic performance coefficients, and part annotations. The dataset encompasses a wide range of geometries and configurations, covering most conventional car design categories and including both detailed underbodies for traditional ICE cars and smooth underbodies for electric cars.

\section{Related Work}
\label{sec:RelatedWork}

Large-scale, diverse, and high-fidelity datasets are essential for advancing deep learning methods for CFD and engineering design, offering standardized data that aid in the development, validation, and comparison of new methodologies. Emerging datasets like AirfRANS~\cite{bonnet2022airfrans}, BubbleML~\cite{hassan2024bubbleml}, Lagrangebench~\cite{toshev2024lagrangebench}, and BLASTNet~\cite{chung2024turbulence} have significantly contributed to the machine learning community in fluid mechanics by providing comprehensive data for training and benchmarking. In engineering design, datasets such as AircraftVerse~\cite{cobb2023aircraftverse} offer detailed and diverse configurations of aerial vehicle designs, helping engineers validate new design strategies and ensure high performance.  However, there still exists no large-scale dataset for 3D shapes that combines both high-fidelity CFD simulations with engineering design specifically tailored for car aerodynamic design.

\begin{table*}[h]
\caption[A comprehensive comparison of various large-scale aerodynamics datasets]{A comprehensive comparison of various aerodynamics datasets, highlighting key aspects such as dataset size, the inclusion of aerodynamic coefficients ($C_d$, $C_l$), velocity ($\textbf{u}$), pressure ($p$), wall-shear stress ($\tau_w$) fields, wheels/underbody modeling, parametric study capability, number of design parameters, shape variation, experimental validation data, multimodality, and open-source availability. \textit{M} refers to 3D meshes, \textit{PC} to point clouds, \textit{P} to parametric data, \textit{A} to part annotations, and \textit{C} to CFD data.\textsuperscript{$\dagger$}Dataset is based on ShapeNet~\cite{chang2015shapenet}. \textsuperscript{*}There exists experimental data from wind tunnel tests for validation.}
\renewcommand{\arraystretch}{1.5}
\setlength{\tabcolsep}{2pt}
\centering
{\scriptsize
\begin{tabular}{cccccccccccccc}
\toprule
\multirow{2}{*}{Dataset} & \multirow{2}{*}{Size} & \multicolumn{5}{c}{Aerodynamics Data} & \multirow{2}{*}{\makecell{Wheels/\\Underbody\\Modeling}} & \multirow{2}{*}{Parametric} & \multirow{2}{*}{\makecell{Design\\Parameters}} & \multirow{2}{*}{\makecell{Shape\\Variation}} & \multirow{2}{*}{\makecell{Experimental\\Validation}} & \multirow{2}{*}{\makecell{Modalities}} & \multirow{2}{*}{\makecell{Open-\\source}} \\
\cmidrule(lr){3-7}
& & $C_d$ & $C_l$ & $\textbf{u}$ & $p$ &  $\tau_w$ & & & & & & & \\ 
\midrule
Usama et al. 2021~\cite{Usama2021} & 500 & \textcolor{mydarkgreen}{\CheckmarkBold} & \textcolor{mydarkred}{\XSolidBrush} & \textcolor{mydarkred}{\XSolidBrush} & \textcolor{mydarkred}{\XSolidBrush} & \textcolor{mydarkred}{\XSolidBrush}  & \textcolor{mydarkred}{\XSolidBrush} & 
\textcolor{mydarkgreen}{\CheckmarkBold} & 40 (2D)  & \textcolor{mydarkred}{\XSolidBrush} & \textcolor{mydarkred}{\XSolidBrush} & \textit{P} & \textcolor{mydarkred}{\XSolidBrush}\\
Li et al. 2023 ~\cite{li2023geometryinformed}\textsuperscript{*} & 551 & \textcolor{mydarkgreen}{\CheckmarkBold} & \textcolor{mydarkred}{\XSolidBrush} & \textcolor{mydarkred}{\XSolidBrush} & \textcolor{mydarkgreen}{\CheckmarkBold} & \textcolor{mydarkgreen}{\CheckmarkBold}  & \textcolor{mydarkred}{\XSolidBrush} & \textcolor{mydarkgreen}{\CheckmarkBold}  & 6 (3D)  & \textcolor{mydarkred}{\XSolidBrush} & \textcolor{mydarkred}{\XSolidBrush}  & \textit{M,P,C} & \textcolor{mydarkred}{\XSolidBrush}\\
Rios et al. 2021~\cite{Rios2021}\textsuperscript{$\dagger$} & 600 & \textcolor{mydarkgreen}{\CheckmarkBold} & \textcolor{mydarkgreen}{\CheckmarkBold} & \textcolor{mydarkred}{\XSolidBrush} &  \textcolor{mydarkred}{\XSolidBrush} & \textcolor{mydarkred}{\XSolidBrush} &  \textcolor{mydarkred}{\XSolidBrush} &    \textcolor{mydarkred}{\XSolidBrush} & - & \textcolor{mydarkred}{\XSolidBrush} &\textcolor{mydarkred}{\XSolidBrush} & \textit{M,PC} & \textcolor{mydarkred}{\XSolidBrush} \\
Li et al. 2023~\cite{li2023geometryinformed}\textsuperscript{$\dagger$}  & 611 & \textcolor{mydarkgreen}{\CheckmarkBold} & \textcolor{mydarkred}{\XSolidBrush} & \textcolor{mydarkred}{\XSolidBrush} & \textcolor{mydarkgreen}{\CheckmarkBold } & \textcolor{mydarkgreen}{\CheckmarkBold} & \textcolor{mydarkred}{\XSolidBrush}  & \textcolor{mydarkred}{\XSolidBrush} & -  & \textcolor{mydarkgreen}{\CheckmarkBold}&\textcolor{mydarkred}{\XSolidBrush}    & \textit{M,C}  & \textcolor{mydarkred}{\XSolidBrush}\\
Umetani et al. 2018~\cite{Umetani2018}\textsuperscript{$\dagger$}& 889 & \textcolor{mydarkgreen}{\CheckmarkBold}& \textcolor{mydarkred}{\XSolidBrush} & \textcolor{mydarkgreen}{\CheckmarkBold} & \textcolor{mydarkgreen}{\CheckmarkBold} & \textcolor{mydarkred}{\XSolidBrush}  & \textcolor{mydarkred}{\XSolidBrush} &   \textcolor{mydarkred}{\XSolidBrush}& -  &\textcolor{mydarkgreen}{\CheckmarkBold}  & \textcolor{mydarkred}{\XSolidBrush} & \textit{M,C} &\textcolor{mydarkgreen}{\CheckmarkBold} \\
Gunpinar et al. 2019~\cite{Gunpinar2019} & 1,000 & \textcolor{mydarkgreen}{\CheckmarkBold} & \textcolor{mydarkred}{\XSolidBrush} & \textcolor{mydarkred}{\XSolidBrush} & \textcolor{mydarkred}{\XSolidBrush}  & \textcolor{mydarkred}{\XSolidBrush} & \textcolor{mydarkred}{\XSolidBrush} &  \textcolor{mydarkgreen}{\CheckmarkBold} & 21 (2D)  & \textcolor{mydarkred}{\XSolidBrush} & \textcolor{mydarkred}{\XSolidBrush} & \textit{P} & \textcolor{mydarkred}{\XSolidBrush}\\
Jacob et al. 2021~\cite{Jacob2021}\textsuperscript{*} & 1,000 & \textcolor{mydarkgreen}{\CheckmarkBold} & \textcolor{mydarkgreen}{\CheckmarkBold} & \textcolor{mydarkgreen}{\CheckmarkBold} &  \textcolor{mydarkred}{\XSolidBrush} & \textcolor{mydarkgreen}{\CheckmarkBold} & \textcolor{mydarkgreen}{\CheckmarkBold} &  \textcolor{mydarkgreen}{\CheckmarkBold} & 15 (3D)  & \textcolor{mydarkred}{\XSolidBrush} & \textcolor{mydarkgreen}{\CheckmarkBold}& \textit{M,C,P} & \textcolor{mydarkred}{\XSolidBrush} \\
Trinh et al. 2024 \cite{trinh20243d}& 1,121 & \textcolor{mydarkred}{\XSolidBrush} & \textcolor{mydarkred}{\XSolidBrush} & \textcolor{mydarkgreen}{\CheckmarkBold} & \textcolor{mydarkgreen}{\CheckmarkBold} & \textcolor{mydarkred}{\XSolidBrush}  & \textcolor{mydarkred}{\XSolidBrush} & \textcolor{mydarkred}{\XSolidBrush} &  -   & \textcolor{mydarkred}{\XSolidBrush} & \textcolor{mydarkred}{\XSolidBrush} & \textit{M,C} & \textcolor{mydarkred}{\XSolidBrush}\\
Remelli et al. 2020~\cite{remelli2020meshsdf}\textsuperscript{$\dagger$} & 1,400 & \textcolor{mydarkred}{\XSolidBrush} & \textcolor{mydarkred}{\XSolidBrush} & \textcolor{mydarkred}{\XSolidBrush} & \textcolor{mydarkgreen}{\CheckmarkBold} & \textcolor{mydarkred}{\XSolidBrush} & \textcolor{mydarkred}{\XSolidBrush} & \textcolor{mydarkred}{\XSolidBrush}  &  -  &\textcolor{mydarkgreen}{\CheckmarkBold} &\textcolor{mydarkred}{\XSolidBrush}  & \textit{M,C} & \textcolor{mydarkred}{\XSolidBrush}\\
Baque et al. 2018~\cite{baque2018geodesic} & 2,000 & \textcolor{mydarkgreen}{\CheckmarkBold} & \textcolor{mydarkred}{\XSolidBrush} & \textcolor{mydarkred}{\XSolidBrush} & \textcolor{mydarkred}{\XSolidBrush}  & \textcolor{mydarkred}{\XSolidBrush} &\textcolor{mydarkred}{\XSolidBrush} &  \textcolor{mydarkgreen}{\CheckmarkBold} & 21 (3D)  & \textcolor{mydarkred}{\XSolidBrush} & \textcolor{mydarkred}{\XSolidBrush} & \textit{M,P} & \textcolor{mydarkred}{\XSolidBrush} \\
Song et al. 2023~\cite{song2023surrogate}\textsuperscript{$\dagger$}  & 2,474 & \textcolor{mydarkgreen}{\CheckmarkBold} & \textcolor{mydarkred}{\XSolidBrush}& \textcolor{mydarkred}{\XSolidBrush} & \textcolor{mydarkred}{\XSolidBrush} & \textcolor{mydarkred}{\XSolidBrush}  & \textcolor{mydarkred}{\XSolidBrush} & \textcolor{mydarkred}{\XSolidBrush} & -  &\textcolor{mydarkgreen}{\CheckmarkBold}   & \textcolor{mydarkred}{\XSolidBrush} & \textit{M} & \textcolor{mydarkgreen}{\CheckmarkBold}\\
DrivAerNet~\cite{elrefaie2024drivaernet}\textsuperscript{*} & 4,000 & \textcolor{mydarkgreen}{\CheckmarkBold} & \textcolor{mydarkgreen}{\CheckmarkBold} & \textcolor{mydarkgreen}{\CheckmarkBold} & \textcolor{mydarkgreen}{\CheckmarkBold} & \textcolor{mydarkgreen}{\CheckmarkBold}  & \textcolor{mydarkgreen}{\CheckmarkBold} & \textcolor{mydarkgreen}{\CheckmarkBold} &  50 (3D)   & \textcolor{mydarkred}{\XSolidBrush} & \textcolor{mydarkgreen}{\CheckmarkBold} & \textit{M,PC,C,P} & \textcolor{mydarkgreen}{\CheckmarkBold} \\
DrivAerNet++ (Ours)\textsuperscript{*} & 8,000 & \textcolor{mydarkgreen}{\CheckmarkBold} & \textcolor{mydarkgreen}{\CheckmarkBold} & \textcolor{mydarkgreen}{\CheckmarkBold} & \textcolor{mydarkgreen}{\CheckmarkBold} & \textcolor{mydarkgreen}{\CheckmarkBold}  & \textcolor{mydarkgreen}{\CheckmarkBold} & \textcolor{mydarkgreen}{\CheckmarkBold}  &  26-50 (3D)   & \textcolor{mydarkgreen}{\CheckmarkBold} & \textcolor{mydarkgreen}{\CheckmarkBold} & \textit{M,PC,C,P,A} & \textcolor{mydarkgreen}{\CheckmarkBold} \\
\bottomrule
\end{tabular}}
\label{tab:datasets_comparison}
\end{table*}

The comparison presented in Table~\ref{tab:datasets_comparison} supports our motivation by highlighting the lack of open-source datasets that encompass a comprehensive range of features for data-driven aerodynamic design. This gap underscores the necessity for datasets that not only provide high-fidelity simulations but also ensure experimental validation to confirm the accuracy and reliability of the computational models. DrivAerNet++ addresses these needs by including multiple data modalities (3D meshes, point clouds, CFD data, parametric data, and part annotations), and considers the modeling of rotating wheels and underbody. While DrivAerNet~\cite{elrefaie2024drivaernet} was based on a single car category, DrivAerNet++ incorporates a variety of car designs and categories.

In the following, we highlight the limitations of the existing datasets:
\begin{itemize}
\item \textbf{Lack of diversity:} The datasets from~\cite{Jacob2021, li2023geometryinformed, elrefaie2024drivaernet, baque2018geodesic, eiximeno2024toward, shen2024car} are based on the same parametric models, resulting in generated cars that stem from the same car designs. This lack of diversity limits the generalization capabilities and creativity in design exploration.
\item \textbf{Small dataset size:} In the engineering design process, changes are not limited to simple geometric parameter adjustments but often involve adding or removing entire components. A significant limitation is the availability of high-quality, watertight meshes necessary for CFD simulations. Most existing datasets~\cite{li2023geometryinformed, Rios2019, Umetani2018, eiximeno2024toward} are either based on ShapeNet~\cite{chang2015shapenet}, which contains very few car designs suitable for CFD and features meshes with low resolutions compared to what is typically used in academia or industry for car aerodynamic design, or are based on morphed geometries from single designs like the Ahmed body~\cite{ahmedbody} or DrivAer\footnote{The DrivAer model~\cite{heft2012introduction, drivAergeometry} is a well-established conventional car reference model developed by researchers at the Technical University of Munich (TUM). It combines features of BMW 3 Series and Audi A4 designs, bridging the gap between simplified models and complex proprietary designs.} body. Consequently, most existing datasets are relatively small, typically on the order of hundreds, with the largest being \cite{song2023surrogate} with 2,474 cars.

\item \textbf{Lower simulation fidelity:} Due to the expensive computational cost of running high-fidelity CFD simulations, there is a trade-off between dataset size and simulation fidelity. As a result, existing datasets, such as \cite{song2023surrogate, baque2018geodesic, remelli2020meshsdf, Rios2019, Usama2021, Gunpinar2019}, used low simulation fidelity, which reduces practical utility.
\end{itemize}
Our dataset, DrivAerNet++, attempts to provide both design variations and diversity, and simulation fidelity, making it highly suitable for conceptual design stages. This balance ensures that designers can explore a wide range of aerodynamic concepts without sacrificing the quality of the simulations.

\section{Dataset Presentation}
\label{sec:DatasetPresentation}
\paragraph{\textbf{Baseline geometry generation}}
In automotive aerodynamics, production cars are typically classified into three categories based on the airflow patterns at their rear end~\cite{heft2012introduction, hucho2013aerodynamik}: estateback, fastback, and notchback cars.  To ensure our dataset covers the entire design space of most conventional car designs, we created multiple parametric models with different designs based on the DrivAer model~\cite{drivAergeometry}. \begin{wrapfigure}{r}{0.5\textwidth} 
    \centering
    \includegraphics[width=0.5\textwidth]{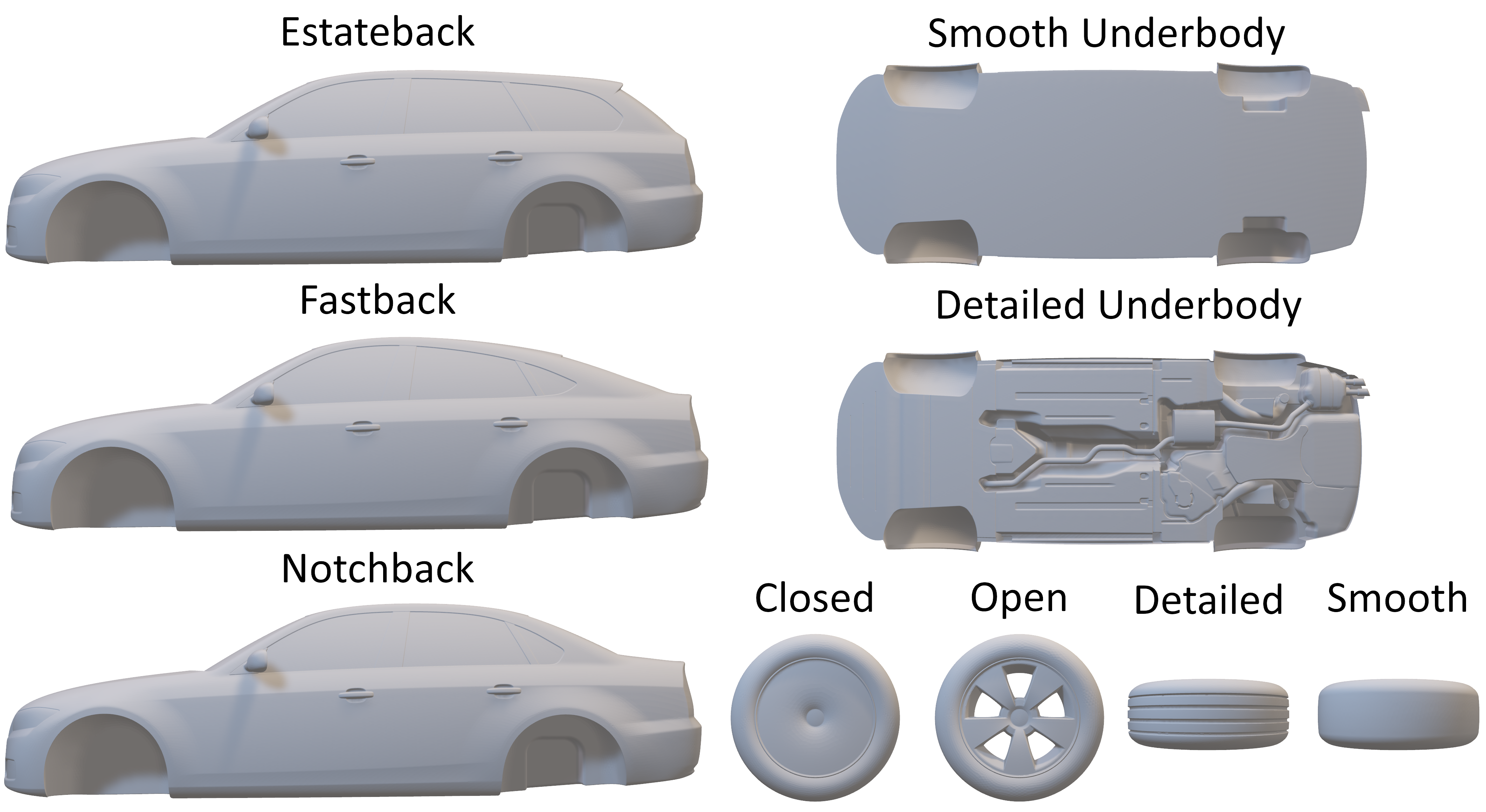}
    \caption{Baseline models from which the parametric models of DrivAerNet++ are derived, demonstrating a range of shape designs and configurations. Variations include estateback, fastback, and notchback car body types alongside different underbody configurations such as smooth and detailed. Wheel options are presented with closed, open, detailed, and smooth styles.}
    \label{fig:DrivAerNet_baselines}
    \vspace{-10pt}
\end{wrapfigure}
This includes various rear configurations—fastback, estateback, and notchback—each leading to different wake structures and flow field patterns. Additionally, we have varied the wheels, incorporating both open and closed designs, as well as smooth and detailed options. For the car underbody, we included both detailed underbodies typical for ICE cars and smooth underbodies suitable for electric cars (see Figure~\ref{fig:DrivAerNet_baselines}).  By exploring various rear, wheels, and underbody configurations, we aim to provide a comprehensive understanding of their aerodynamic impacts, thus supporting the development of more robust and generalizable deep learning models.
For the creation of the parametric models, we utilized the commercial software ANSA® to define 26 geometric parameters, allowing us to morph these parametric models, resulting in a large-scale dataset of 3D cars. 
Our goal was to develop a procedural generator that creates topologically valid car designs, ensuring each design meets the necessary requirements for evaluation by CFD solvers and usability by automotive designers.

\begin{figure}[h]
    \centering
    \includegraphics[width=\textwidth]{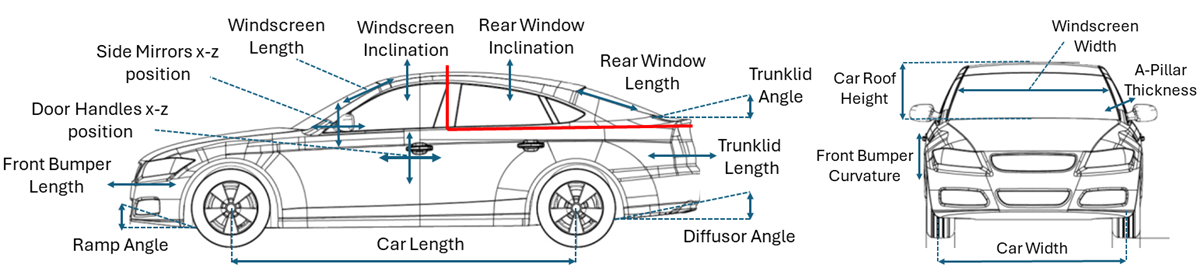}
\caption{Design parameters for the generation of the DrivAerNet++ dataset. Several geometric parameters with significant impact on aerodynamics were selected and varied within a specific range. These parameter ranges were chosen to avoid values that are either difficult to manufacture or not aesthetically pleasing. The car sketch is adapted from \cite{heft2012experimental}.}
    \label{fig:DrivAerNet_DesignParams}
\end{figure}

\paragraph{\textbf{High-resolution 3D industry-standard designs}}
Our design curation involved selecting valid car configurations followed by detailed CFD simulations to evaluate aerodynamic performance. We aimed to create a balanced dataset that encompasses a wide variety of car designs, ensuring coverage of diverse aerodynamic performance metrics and aesthetic considerations. Figure~\ref{fig:DrivAerNet_DesignParams} shows a subset of the design parameters used for the generation of the DrivAerNet++ dataset. By defining a lower and upper bound for each parameter and morphing the baseline parametric model, we ensure a comprehensive and well-defined representation suitable for engineering applications and simulations. Our design methodology is diversity-preserving, ensuring that optimization does not result in overly similar designs. To achieve this, we employed Optimal Latin Hypercube sampling for the Design of Experiments (DoE). Specifically, we used the Enhanced Stochastic Evolutionary Algorithm (ESE)~\cite{damblin2013numerical} to ensure efficient sampling of the design space. Using these steps, DrivAerNet++ is significantly more diverse than DrivAerNet~\cite{elrefaie2024drivaernet}.

\paragraph{Comparison of mesh resolutions across large-scale datasets}

Here we compare the mesh resolution on car surfaces across different existing large-scale datasets. The results are summarized in Table~\ref{tab:mesh_comparison}. The mesh resolution comparison highlights the varying levels of detail used in different large-scale datasets for aerodynamic simulations. DrivAerNet++ (ours) provides the highest mesh resolution, ranging from 350,000 to 750,000 cells, ensuring detailed representation of car surfaces for accurate simulations. In contrast, other datasets like the Ahmed body by Li et al. (2023)~\cite{li2023geometryinformed} and ShapeNet (Song et al., 2023~\cite{song2023surrogate}) use lower resolutions, with 100,000 cells and 30,000--50,000 cells, respectively. This finer mesh resolution in DrivAerNet++ enables better fidelity in aerodynamic analyses, particularly for capturing intricate details crucial for engineering design.

\begin{table}[h]
\centering
\scriptsize
\caption{Comparison of mesh resolutions on car surfaces in large-scale datasets.}
\label{tab:mesh_comparison}
\vspace{3pt}
\begin{tabular}{lc}
\hline
Dataset & Mesh Resolution on Car Surface \\ \hline
DrivAerNet++ (Ours) & 350k to 750k cells \\ 
Ahmed body (Li et al., 2023)~\cite{li2023geometryinformed} & 100k cells \\ 
ShapeNet (Song et al., 2023)~\cite{song2023surrogate} & 30--50k cells \\ 
ShapeNet (Umetani et al., 2018)~\cite{Umetani2018} &  10k cells \\ \hline
\end{tabular}
\end{table}

\paragraph{\textbf{CFD mesh generation}}

For the mesh generation, we utilized the open-source SnappyHexMesh utility~\cite{openfoam2023snappyhexmesh}. Following the best meshing practices~\cite{heft2011investigation, QIN2024105711, heft2012experimental}, we ensured that our meshes accurately model boundary layer interactions. Each mesh consists of 24 million cells in total, with 500-750k cells specifically dedicated to the car surface, ensuring detailed meshing for the car body and wheels to accurately capture the necessary aerodynamic phenomena. For comparison, the largest dataset after DrivAerNet~\cite{elrefaie2024drivaernet} and DrivAerNet++, introduced by~\cite{song2023surrogate}, which has 2474 car designs, used about 2 million cells for each CFD simulation.  All technical details regarding the meshing process and validation are provided in the supplementary.

\paragraph{\textbf{Automated high-fidelity CFD simulations}}

We employed the open-source software OpenFOAM® v11~\cite{OpenFOAMv11} to conduct steady-state incompressible simulations using the $k$-$\omega$ SST turbulence model, based on Menter’s formulation~\cite{menter2003ten}. We ran quality checks for the generated geometries to ensure they were simulation-ready and correctly aligned within the CFD domain, followed by quality checks for the CFD meshing, and finally, checks to ensure the convergence of each CFD simulation. In total, we generated an additional 4,000 simulations to our recently published DrivAerNet dataset~\cite{elrefaie2024drivaernet}, encompassing various designs, flow behaviors, turbulence, and separation phenomena.

\paragraph{\textbf{Computational cost}}
Running the high-fidelity CFD simulations for DrivAerNet++ required substantial computational resources. The simulations were conducted on the MIT Supercloud, leveraging parallelization across 60 nodes, totaling 2880 CPU cores, with each CFD case using 256 cores and 1000 GBs of memory. The full dataset requires \textbf{39 TB} of storage space, and the total number of files for the CFD simulations amounts to 834,332. The job parallelization was managed using MPI~\cite{openmpi_mpirun}, ensuring efficient distribution of computational tasks. The simulations took approximately \textbf{3 $\times$ 10$^{\mathbf{6}}$ CPU-hours} to complete.

\paragraph{\textbf{Dataset structure}}

Our dataset represents car designs using multiple modalities to ensure comprehensive coverage and ease of use in various applications.
\begin{itemize}
    \item 3D Car Designs: We provide 3D STL meshes ideal for engineering design, CFD analysis, design optimization, and generative AI applications.
\item ANSA® 3D Parametric Models: Provided to enable researchers to generate their own datasets and incorporate custom parameters as needed for specific research objectives.
    \item Tabular Parametric Data: Each 3D car geometry is parametrized with 26 parameters that completely describe the design. This data can be used for parametric regression, classification, and design feature importance analysis.
    \item Aerodynamic Performance Data: Includes force coefficients such as drag $C_d$, lift (total $C_l$, rear $C_{l,r}$, front $C_{l,f}$), and moment $C_m$ values, both mean and standard deviation.
    \item CFD Data: Includes raw and post-processed data, 3D full flow field information with velocity and pressure, surface fields with pressure and wall shear stress, streamlines around the car body, and 2D slices.
    \item Point Cloud: Representations uniformly sampled over the car surface in 5k, 10k, 100k, 250k, and 500k nodes.
    \item Annotated Labels: For both car designs and segmentation of different car components of each individual car, useful for tasks such as object detection, semantic segmentation, parametric studies, and automated CFD meshing.
\end{itemize}

\begin{figure}[h]
    \centering
    \includegraphics[width=\textwidth]{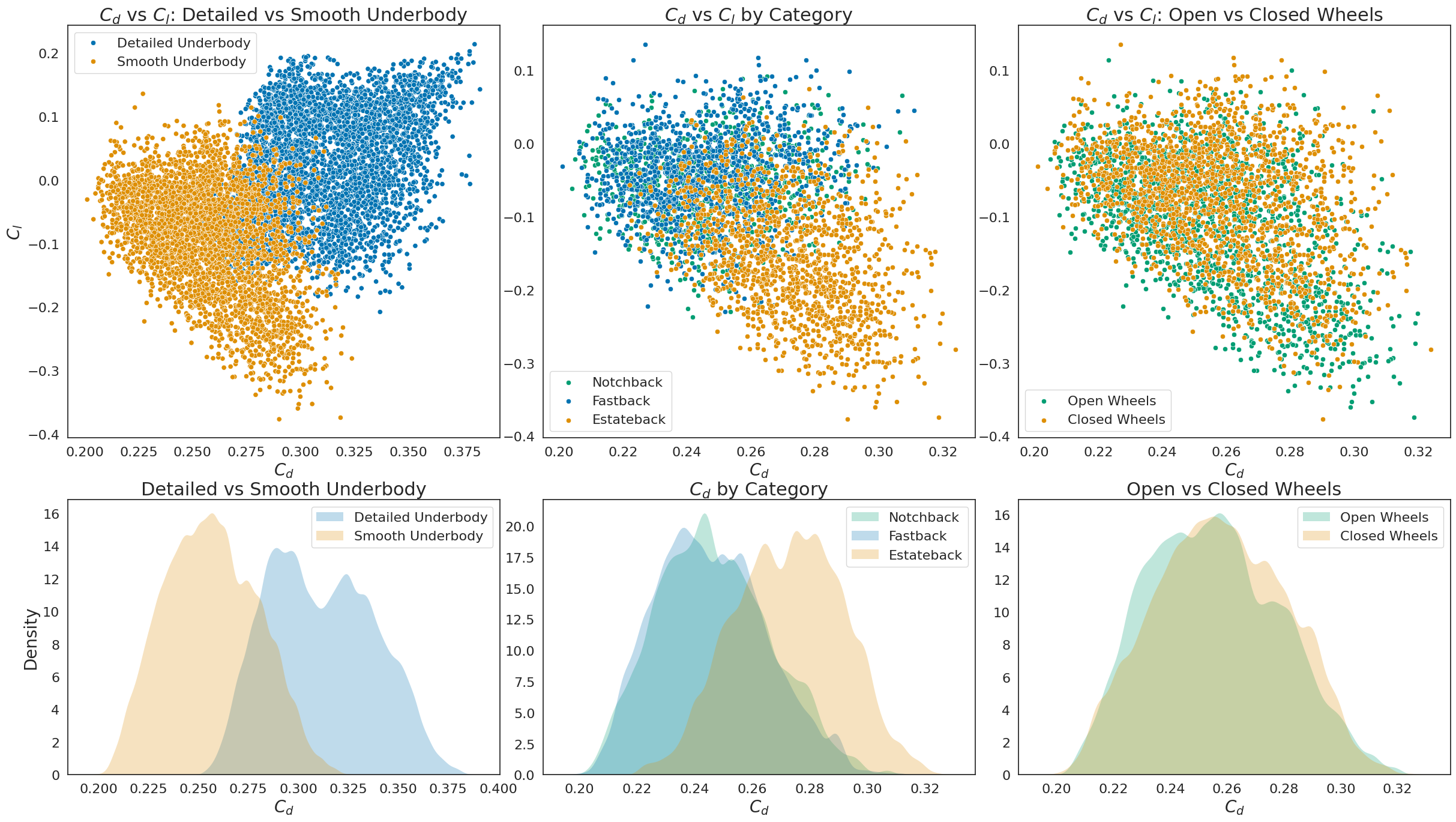}
\caption{The scatter plots in the top row illustrate the relationship between $C_d$ and $C_l$ for different configurations: the first plot shows the influence of underbody configurations, comparing detailed versus smooth underbodies typically used in electric cars. The second plot highlights the impact of design aesthetics and style across car categories (notchback, fastback, and estateback). The third plot examines the effect of different wheel configurations, emphasizing the significance of small geometric modifications on aerodynamics. The density plots in the bottom row show the distribution of $C_d$ for the same configurations, providing a detailed view of how these design elements and categories influence aerodynamic efficiency.}
  \label{fig:DrivAerNet_scatter_kde}
\end{figure}

Figure \ref{fig:DrivAerNet_scatter_kde} shows a comparative analysis of aerodynamic performance across various design configurations and categories, highlighting the diversity and size of our dataset.

We provide detailed metadata for the DrivAerNet++ dataset using the Croissant format~\cite{mlcroissant} to ensure comprehensive documentation and ease of use for the research community. We also include datasheets for datasets~\cite{gebru2021datasheets}, and the dataset is provided under the Creative Commons Attribution-NonCommercial (CC BY-NC) license. DrivAerNet++ will be hosted on the Harvard Dataverse Repository to ensure optimal accessibility and systematic data management. Since the 39 TB of data might pose challenges for data sharing and access, we also provide subsets of our dataset tailored for different tasks, with detailed metadata included to facilitate usability.

\paragraph{Scaling up the dataset size vs simulation fidelity}
Scaling the dataset size in engineering design and CFD poses significant challenges due to the high computational and storage requirements. Despite these challenges, we argue that our simulation fidelity, mesh size, and the use of the \(k\)-\(\omega\) SST turbulence model are sufficient for the initial design stages.  The results shown in Section~\ref{subsubsection:parametric_data} and in our previous work~\cite{elrefaie2024drivaernet} highlight the importance of scaling the training dataset size for better generalization to unseen new designs. While hybrid RANS-LES methods with mesh sizes of \(\mathcal{O}(100M)\)~\cite{ashton2016assessment, ashton2023summary,eiximeno2024toward} cells provide more accurate turbulence modeling, their computational cost and storage demands are prohibitive. The \(k\)-\(\omega\) SST model offers a balanced approach that supports the scaling of training dataset sizes for deep learning models. Our dataset surpasses all open-source datasets and most literature in terms of simulation fidelity and size, enabling good coverage of the design space and providing insights into various design strategies.

Leveraging multi-fidelity data and transfer learning for efficient surrogate model development has shown promising results, as demonstrated by~\cite{elrefaie2024real} for flow field prediction and by~\cite{tao2024multi} for multi-fidelity shape optimization. Utilizing large-scale and diverse datasets like DrivAerNet++ facilitates a comprehensive examination of the design space. Subsequent fine-tuning on higher fidelity datasets from hybrid RANS-LES simulations can further improve predictive capabilities. This approach mirrors strategies in the broader deep learning community, demonstrating the synergy between data scale and model refinement for significant performance enhancements.

\section{Benchmarking Setup}
In this paper, we explore various machine learning tasks, specifically focusing on surrogate modeling (regression) of aerodynamic drag (\(C_d\)). This study is distinctive as it is the first to benchmark diverse models using a large-scale, diverse, and high-fidelity dataset. While previous research~\cite{elrefaie2024drivaernet,Jacob2021,li2023geometryinformed,baque2018geodesic,Umetani2018,Usama2021} typically limits comparisons to a single car design or category, our approach, in contrast, enables fair and generalized comparisons across models by employing a comprehensive and publicly accessible dataset, showcasing real-world applicability and performance in automotive aerodynamics.

\paragraph{\textbf{Metrics and visualization}}
We assess different models' performance using several metrics: Mean Squared Error (MSE) quantifies the average squared differences between predicted and actual values, making it sensitive to large errors. Mean Absolute Error (MAE) measures the average magnitude of errors and is less affected by outliers. Maximum Absolute Error (Max AE) identifies the largest prediction error, indicating worst-case accuracy. The Coefficient of Determination (\(R^2\) Score) represents the proportion of variance explained by the model, with a value of 1 indicating a perfect fit. Lower MSE and MAE values, along with a higher \(R^2\) score and a lower Max AE, indicate more accurate predictions. For \(C_d\) estimation, an MAE of less than 0.005 compared to wind tunnel measurements is considered acceptable~\cite{schutz2013hucho, Jacob2021}.

\section{Benchmarking Results}

\subsection{Surrogate modeling of the aerodynamic drag}
In the conceptual and initial phases of car design, the aerodynamic drag coefficient is a critical metric as it indicates design efficiency and impacts driving range. Therefore, having accurate and rapid drag estimates is crucial in the design process. In this section, we introduce two approaches for aerodynamic drag prediction: first, using 3D geometric deep learning with 3D meshes, and second, using automated machine learning based on the parametric dataset.

\subsubsection{Aerodynamic drag prediction 
 from 3D meshes using geometric deep learning}

Here, we test different geometric deep learning models (PointNet~\cite{qi2017pointnet}, GCNN~\cite{kipf2016semi}, and RegDGCNN~\cite{elrefaie2024drivaernet}) implemented in PyTorch~\cite{paszke2019pytorch} and PyTorch Geometric~\cite{fey2019fast} for the task of surrogate modeling of aerodynamic drag, highlighting the importance of dataset diversity and scaling.
Specifically, we train models using different representations, including graph-based and point cloud-based models. 
\begin{table*}[h]
\caption{Comparative analysis of deep learning models for aerodynamic drag prediction on the test set from the DrivAerNet~\cite{elrefaie2024drivaernet} dataset (fastback design with open wheels, with mirrors, and with detailed underbody configuration) comprising 600 car designs.}
\scriptsize
\centering
\setlength{\tabcolsep}{3pt} 
\begin{tabular}{cccccccc}
\hline
\textbf{Model} & \textbf{MSE  $\downarrow$} & \textbf{MAE $\downarrow$} & \textbf{Max AE $\downarrow$} & \textbf{$R^2$ $\uparrow$} & \textbf{Training Time $\downarrow$} & \textbf{Inference Time $\downarrow$} & \textbf{Number of Parameters} \\
& $(\times 10^{-5})$ & $(\times 10^{-3})$ & $(\times 10^{-3})$ & & & & \\
\hline
PointNet~\cite{qi2017pointnet} & 12.0  & 8.85 & 10.18  & 0.826  & 0.5hrs & 0.51s  &  2,348,097 \\
GCNN~\cite{kipf2016semi}& 10.7 & 7.17 & 10.97 & 0.874 & 10.4hrs & 20.71s & 100,481 \\ 
RegDGCNN~\cite{elrefaie2024drivaernet} & 8.01 & 6.91  & 8.80  & 0.901  & 3.2hrs  & 0.52s & 3,164,257 \\
\hline
\end{tabular}
\label{tab:model_comparison_1}
\end{table*}
A different approach from previous studies~\cite{Jacob2021, eiximeno2024toward, li2023geometryinformed, baque2018geodesic} is taken by training the models on a single car design (fastback) and, in another experiment, on all designs (fastback, notchback, and estateback).
First, we train the deep learning models on the DrivAerNet dataset~\cite{elrefaie2024drivaernet}, which includes variants of the fastback with detailed underbody, open wheels, and mirrors. This dataset comprises 4,000 car designs (2,800 for training, approximately 600 for validation, and 600 for testing), with results presented in Table~\ref{tab:model_comparison_1}. Then, we train and test the same models on the DrivAerNet++ dataset, which has 8,000 car designs with extensive variations (fastback, estateback, notchback, smooth and detailed underbodies, and different wheel configurations), divided into 5,600 for training, 1,200 for validation, and 1,200 for testing, with results shown in Table~\ref{tab:model_comparison_2}.

The shape variation in DrivAerNet++ (refer to Figure~\ref{fig:DrivAerNet_scatter_kde}) introduces additional challenges. For example, replacing the detailed underbody with a smooth underbody can shift the drag distribution for the same car design. Changing the wheels from open to closed can slightly affect the drag. Additionally, different rear configurations can result in varying flow field separation behaviors, causing significant changes in drag values. These factors make DrivAerNet++ a very challenging task for generalization, as seemingly minor changes can significantly impact drag values and present difficulties for deep learning models in accurately learning the features of these variations.

\begin{table*}[ht]
\caption{Comparative analysis of deep learning models for aerodynamic drag prediction on the test set from DrivAerNet++ (All cars) comprising 1,200 car designs.}
\scriptsize
\centering
\setlength{\tabcolsep}{3pt} 
\begin{tabular}{cccccccc}
\hline
\textbf{Model} & \textbf{MSE  $\downarrow$} & \textbf{MAE $\downarrow$} & \textbf{Max AE $\downarrow$} & \textbf{$R^2$ $\uparrow$} & \textbf{Training Time $\downarrow$} & \textbf{Inference Time $\downarrow$}  & \textbf{Number of Parameters} \\
& $(\times 10^{-5})$ & $(\times 10^{-3})$ & $(\times 10^{-3})$ & & & & \\
\hline
PointNet~\cite{qi2017pointnet} & 14.9 & 9.60 & 12.45 & 0.643 &  2.06hrs & 0.84s &  2,348,097 \\
GCNN~\cite{kipf2016semi}& 17.1 & 10.43 & 15.03 & 0.596 & 49hrs & 50.8s & 100,481 \\
RegDGCNN~\cite{elrefaie2024drivaernet} & 14.2 & 9.31 & 12.79 & 0.641 & 12.6hrs & 0.85s & 3,164,257 \\
\hline
\end{tabular}
\label{tab:model_comparison_2}
\end{table*}

\subsubsection{Aerodynamic drag prediction from tabular parametric data}
\label{subsubsection:parametric_data}
We also approach the task of aerodynamic drag prediction using parametric data. For this purpose, we employ an AutoML (automated machine learning) framework~\cite{erickson2020autogluon} optimized with Bayesian hyperparameter tuning, along with models such as Gradient Boosting~\cite{friedman2001gradientboosting}, XGBoost~\cite{chen2016xgboost}, LightGBM~\cite{ke2017lightgbm}, and Random Forests~\cite{breiman2001randomforest}. These methods leverage design parameters to estimate aerodynamic drag, eliminating the need for detailed 3D geometry. Such an approach is invaluable for efficiently evaluating the impact of geometric modifications on drag and overall car performance. The use of parametric data offers significant advantages due to its accessibility and ease of manipulation compared to 3D mesh modifications. Engineers can swiftly adjust design parameters and immediately observe the effects on aerodynamic performance, thereby streamlining the design process.

We train the models on a single parametric car design (fastback) and, in another experiment, on all parametric models (fastback, notchback, and estateback), to explore how well the models generalize across different designs. For both experiments, we split the dataset into 80$\%$ for training and 20$\%$ for testing. We then further divided the training set into subsets at 20$\%$, 40$\%$, 60$\%$, 80$\%$, and 100$\%$ of the training portion. 
\begin{figure}[h!]
    \centering
    \includegraphics[width=\textwidth]{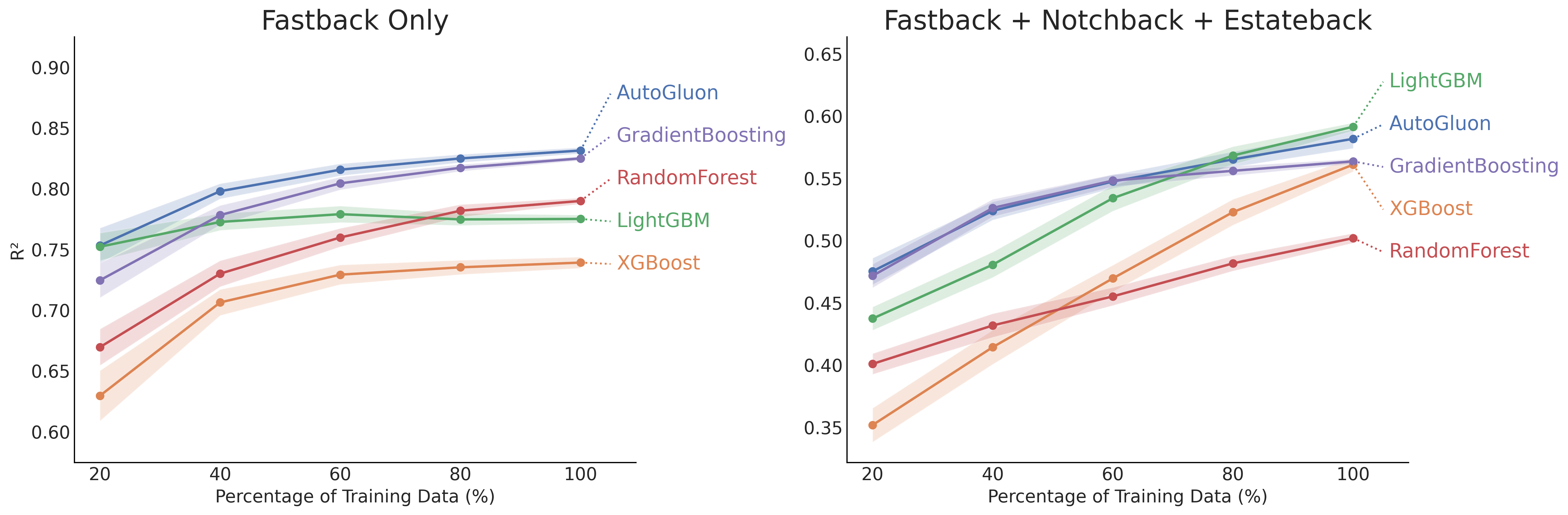}
    \caption{Drag coefficient prediction based on the parametric data for different car categories. The plots show the median and 95\% confidence interval of the $R^2$ score as a function of the percentage of the training data.}
    \label{fig:AutoML_drag_coefficient_prediction}
\end{figure}

To standardize the parametric studies, we focused on 26 parameters instead of 50, as the 50 geometric parameters model from DrivAerNet~\cite{elrefaie2024drivaernet} was based only on one car category, specifically the fastback with a detailed underbody and open wheels. The results, illustrated in Figure~\ref{fig:AutoML_drag_coefficient_prediction}, reveal that AutoGluon performs better in the single fastback category, whereas LightGBM excels on the combined dataset. Nonetheless, performance declines for all models on the combined dataset. A significant finding is that for all models, both in single and combined categories, enlarging the dataset size leads to enhanced performance. For example, the \( R^2 \) value for XGBoost increases from approximately 0.35 to 0.55 by increasing the training set size from 640 to 3,200.

\section{Conclusion}In this paper, we introduced DrivAerNet++, the largest, multimodal 3D dataset for data-driven aerodynamic design, incorporating high-fidelity CFD simulations and a diverse array of car designs. Our dataset includes 8,000 cars based on industry-standard shapes, offering extensive coverage of various aerodynamic performance metrics. The dataset, requiring 39 TB of storage, is significantly larger than comparable datasets in engineering and is made publicly available. The computational cost of generating DrivAerNet++ was an order of magnitude larger than that of the recently published CFD dataset from \cite{li2023geometryinformed}, which utilized 185,744 CPU-hours, whereas ours required 3,000,000 CPU-hours.

This comprehensive dataset supports a wide range of machine learning tasks, including surrogate modeling of aerodynamic performance, acceleration of CFD simulations, data-driven design optimization, generative AI, shape and part classification, and 3D shape reconstruction. We also presented the first benchmarking results, demonstrating the effectiveness of geometric deep learning models and AutoML frameworks for drag coefficient prediction. Furthermore, we explored the challenges of creating a generalized model for surrogate modeling of drag across different car categories. While models trained on a single car category performed well, their performance in terms of \( R^2 \) decreased significantly from 0.82-0.9 to 0.6 when applied to the full diverse dataset. This underscores the complexity of achieving robust performance across varied designs.

Our dataset can be utilized for data-driven design of both internal combustion engine (ICE) cars and electric vehicles, covering main design aspects such as aesthetics/style and aerodynamic efficiency and performance. We believe this dataset will serve as a cornerstone for advancing research in engineering design and CFD, offering rich resources for developing more accurate and efficient predictive models.

\section{Limitations and Future Work}
\label{sec:Limitations_FutureWork}

The surrogate models we trained on DrivAerNet++ are currently not complex enough to fully capture the intricate geometric and aerodynamic features present in diverse car designs. More advanced models, such as Geometry-Informed Neural Operators~\cite{li2023geometryinformed}, Convolutional Occupancy Networks~\cite{peng2020convolutional}, and sophisticated graph models, should be explored to improve predictive performance. We primarily focused on the surrogate modeling of drag, as it is a critical factor in the initial design phase. However, leveraging DrivAerNet++ for additional applications, such as accelerating CFD simulations, would provide a more comprehensive tool for aerodynamic design.

To enhance the dataset, future work will focus on integrating transient CFD simulations and incorporating additional modalities, such as 2D image renderings and multimodal learning approaches. These enhancements aim to improve model accuracy and robustness, fostering further innovation in automotive design and optimization.

\begin{ack}
The authors thank Ravi Nimbalkar and Yatin Kumbhar from BETA CAE Systems USA Inc for their support in creating the parametric models in ANSA®. The authors acknowledge the MIT SuperCloud and Lincoln Laboratory Supercomputing Center for providing HPC and consultation resources that have contributed to the research results reported within this paper. Special thanks to Chris Hill for providing access to MIT SuperCloud and to Lauren Milechin for her support in setting up the simulation environment. Additionally, the authors wish to thank Amy Nurnberger from MIT and Sonia Barbosa from Harvard University for their support with uploading the dataset to Harvard Dataverse. Finally, the authors extend their gratitude to Leonid Andreev for his technical support with uploading the dataset to Harvard Dataverse.
\end{ack}

\newpage

\section*{Checklist}

\begin{enumerate}

\item For all authors...
\begin{enumerate}
  \item Do the main claims made in the abstract and introduction accurately reflect the paper's contributions and scope?
    \answerYes{}
  \item Did you describe the limitations of your work?
    \answerYes{} This is done in Section~\ref{sec:Limitations_FutureWork} 
  \item Did you discuss any potential negative societal impacts of your work?
    \answerNA{}
  \item Have you read the ethics review guidelines and ensured that your paper conforms to them?
    \answerYes{} Our work does not involve human participants or crowd-sourcing. The data used is licensed, consented, and adheres to the FAIR (Findable, Accessible, Interoperable, and Reusable) principles~\cite{wilkinson2016fair}.

\end{enumerate}

\item If you are including theoretical results...
\begin{enumerate}
  \item Did you state the full set of assumptions of all theoretical results?
    \answerNA{}
	\item Did you include complete proofs of all theoretical results?
    \answerNA{}
\end{enumerate}

\item If you ran experiments (e.g. for benchmarks)...
\begin{enumerate}
  \item Did you include the code, data, and instructions needed to reproduce the main experimental results (either in the supplemental material or as a URL)?
    \answerYes{} The source code utilized for model training and assessment in this research can be accessed at the given repository \url{https://github.com/Mohamedelrefaie/DrivAerNet}, where the README file offers detailed guidance for usage.
  \item Did you specify all the training details (e.g., data splits, hyperparameters, how they were chosen)?
    \answerYes{} DrivAerNet++ is designed as a benchmark dataset for 3D problems, complete with a split into training, validation, and test sets to ensure fair comparisons among various machine and deep learning models. Details on the dataset splits can be found here: \url{https://github.com/Mohamedelrefaie/DrivAerNet}
	\item Did you report error bars (e.g., with respect to the random seed after running experiments multiple times)?
    \answerNo{}
	\item Did you include the total amount of compute and the type of resources used (e.g., type of GPUs, internal cluster, or cloud provider)?
    \answerYes{} The high-fidelity CFD simulations for DrivAerNet++ were conducted on the MIT SuperCloud, utilizing parallelization across 60 nodes, totaling 2880 CPU cores. The simulations took approximately 3 × 10$^6$ CPU-hours to complete. For more details, refer to Section \ref{sec:DatasetPresentation}: Computational cost.
\end{enumerate}

\item If you are using existing assets (e.g., code, data, models) or curating/releasing new assets...
\begin{enumerate}
\item If your work uses existing assets, did you cite the creators?
\answerYes{} The baseline CAD geometry of the DrivAer model is available online for free download~\cite{drivAergeometry}, but registration is required, and we have cited the creators.
\item Did you mention the license of the assets?
\answerNA{} The CAD geometry is not provided under a license, however it requires registration for access.
\item Did you include any new assets either in the supplemental material or as a URL?
\answerYes{} All new car designs created by the authors of this paper are included in the dataset and will be hosted on Harvard Dataverse Repository.
\item Did you discuss whether and how consent was obtained from people whose data you're using/curating?
\answerYes{} Consent is not applicable as all car designs were created by the authors of this paper.
\item Did you discuss whether the data you are using/curating contains personally identifiable information or offensive content?
\answerNA{} The data does not contain any personally identifiable information or offensive content.
\end{enumerate}

\item If you used crowdsourcing or conducted research with human subjects...
\begin{enumerate}
  \item Did you include the full text of instructions given to participants and screenshots, if applicable?
      \answerNA{}
  \item Did you describe any potential participant risks, with links to Institutional Review Board (IRB) approvals, if applicable?
      \answerNA{}
  \item Did you include the estimated hourly wage paid to participants and the total amount spent on participant compensation?
       \answerNA{}
\end{enumerate}

\end{enumerate}


\appendix

\newpage
\startcontents[appendices]  
\phantomsection

\section*{Table of Contents for Appendices}
\printcontents[appendices]{}{1}{}  

\newpage

\section{Reproducibility Statement}
We provide a GitHub repository to facilitate the reproduction of all experiments in the main paper, along with links to download the dataset. The repository includes the following: code to replicate the machine learning experiments and code to generate the figures presented in the paper.  The full dataset will be made available after the peer-review process is complete. The training and experiments were performed on a machine equipped with AMD EPYC 7763 64-Core Processors, totaling 256 CPU cores, and four Nvidia A100 80GB GPUs. 
\section{URL and Links}
\textbf{DrivAerNet++ Dataset: } The DrivAerNet~\cite{elrefaie2024drivaernet} and the DrivAerNet++ datasets are available on our GitHub repository at \url{https://github.com/Mohamedelrefaie/DrivAerNet}.
\newline
\newline
\textbf{Croissant Metadata:} The URL to Croissant metadata record documenting the dataset/benchmark is available on our GitHub repository at \url{https://github.com/Mohamedelrefaie/DrivAerNet/tree/main/mlcroissant}.
\newline
\newline
\textbf{DL Code:} The source code utilized for model training and assessment in this research can be accessed at the given repository \url{https://github.com/Mohamedelrefaie/DrivAerNet}, where the README file offers detailed guidance for usage.
\newline
\newline
\textbf{Tutorials and Browsing Tools:} Additional resources, including data handling tutorials and contribution guidelines, are available at \url{https://github.com/Mohamedelrefaie/DrivAerNet/tree/main/tutorials}

\section{Author Statement and Data License Confirmation}
We, the authors, hereby declare that we bear full responsibility for any violations of rights or other issues arising from the use of the data. We confirm that the dataset is available under the Creative Commons Attribution-NonCommercial (CC BY-NC) license. This license allows others to remix, tweak, and build upon our work non-commercially, and although their new works must also acknowledge our original work and be non-commercial, they don’t have to license their derivative works on the same terms.

\newpage
\section{Broader Impact}

The DrivAerNet++ dataset can be used for several tasks, such as data-driven design optimization, training surrogate models, data-driven turbulence modeling, accelerating CFD, 3D shape and part classification, 3D shape reconstruction, and generative design. 
We are confident that DrivAerNet++ will serve as a highly valuable resource for multiple research communities in various ways:

\begin{enumerate}
\item \textbf{Advancing Automotive Aerodynamic Design and Accelerating Design Cycles:} Enabling detailed studies in aerodynamics to foster innovation in car design for enhanced performance and efficiency, while assisting CFD engineers in creating more efficient and accurate surrogate models, speeding up design iteration cycles, and democratizing simulations to provide designers with greater freedom.
  \item \textbf{Accelerating CFD Simulations:} DrivAerNet++ provides high-fidelity simulation data that can be used to train surrogate models, reducing the computational cost and time required for CFD simulations. This allows for faster iterations and more efficient exploration of the design space.
  \item \textbf{Machine Learning Integration:} Offering a rich resource for training and testing advanced machine learning models, including those in geometric deep learning, to understand complex aerodynamic phenomena.
  \item \textbf{Benchmarking and Validation:} Serving as a benchmark dataset for ML algorithms, aiding in the validation and comparison of different modeling approaches.

\item \textbf{Environmental Impact:} Contributing to the development of more aerodynamically efficient cars, which can lead to reduced fuel consumption and lower emissions, and extending the range of electric cars.

  \item \textbf{Safety and Prototyping:} Reducing the need for physical prototyping and wind tunnel testing, thereby lowering costs and enhancing safety by enabling virtual testing of numerous design variants.
  \item \textbf{Cross-Disciplinary Insights:} Offering insights into the intersection of CFD, ML, and automotive design, encouraging cross-disciplinary research and collaboration.
\end{enumerate}

This multimodal, large-scale, and high-fidelity CFD dataset of 8000 car shapes is expected to catalyze significant advancements in both the academic and industrial realms. The methodology and approach implemented here are not only limited to car design but can also be extended to other applications involving numerical simulations and deep learning, such as aerospace design.

\newpage

\section{DrivAerNet++ Dataset Generation}
\subsection{Generating a wide range of car designs}
The goal is to create a large-scale dataset of 3D shapes of cars with high-quality mesh resolutions, watertight geometries, and diverse configurations that resemble real-world car designs. The ShapeNet~\cite{chang2015shapenet} dataset offers a wide range of car shapes; however, as discussed, the mesh resolution and non-watertightness make it unsuitable for large-scale engineering designs or CFD datasets. Other generic car models, like the Ahmed body~\cite{ahmedbody} or the Windsor body~\cite{windsorbody} are very simple and do not represent actual cars with realistic shapes. Therefore, we chose the DrivAer body~\cite{heft2012introduction,drivAergeometry} and parametrized the model with different configurations. In addition, the choice of the DrivAer model is supported by the availability of both computational and experimental references, allowing us to validate our results against established data~\cite{heft2012introduction,heft2012experimental,heft2011investigation}.

\begin{figure}[h]
    \centering
\includegraphics[width=\linewidth]{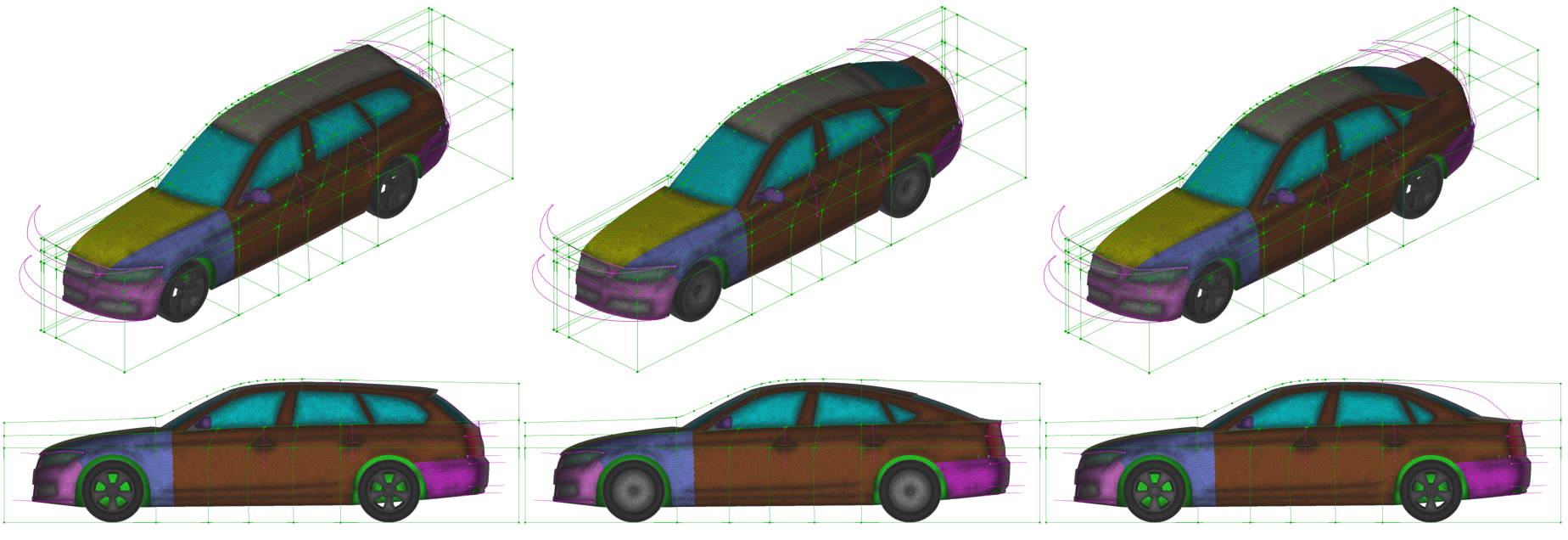}
    \caption{Three parametric models from DrivAerNet++. The green boxes represent the morphing boxes, and the purple curves illustrate the trajectories followed during the morphing of the car geometry. A total of 26 parameters were utilized to adjust and refine the car designs.}
    \label{fig:ANSA_parametricModels}
\end{figure}

The parametric models\footnote{Available at: \url{https://github.com/Mohamedelrefaie/DrivAerNet/ParametricModels}}, along with the constraints and the bounds applied during the Design of Experiments (DoE) were generated using the commercial software ANSA® (see Figure~\ref{fig:ANSA_parametricModels}).
We employed two of the ANSA® advanced tools to create the parametric models and generate the DoE: the morphing tool and the optimization tool. The morphing tool allows for versatile processing of alternative variations in each model, with functions that can reshape geometry entities. To create diverse car designs, we used two main morphing methods: morphing boxes and direct morphing.
The morphing boxes can load any type of entities; in our case, we used this method to create four parameters that control the overall length, width, roof height, and greenhouse angle of the car. 

To create detailed parameters that control specific features of the car model, such as windshield inclination, ramp angle, diffusor angle, etc., the direct morphing method was employed using different movement types provided by the DFM (Direct Fitting Movements) function.

\begin{figure}[h]
    \centering
    \includegraphics[width=\textwidth]{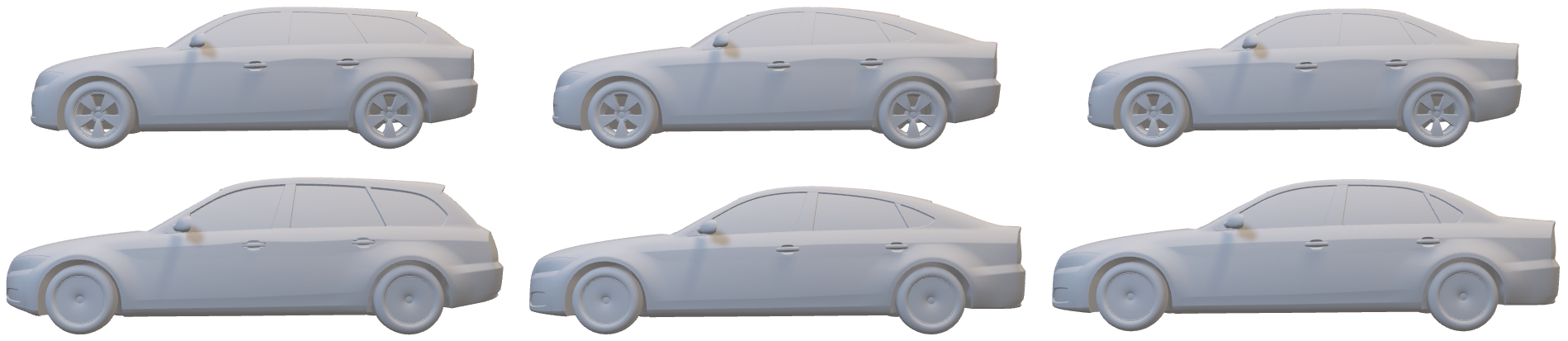}
    \caption{Variations of estateback, fastback, and notchback car designs from left to right, showcasing models with the smallest (top) to largest (bottom) volume.}
    \label{fig:DrivAerNet_Volumes}
\end{figure}
Figure~\ref{fig:DrivAerNet_Volumes} illustrates the variations in car designs, highlighting the range from smallest to largest volumes across estateback, fastback, and notchback models. This illustration highlights the extensive range covered by the DrivAerNet++, which includes all conventional production car configurations to ensure comprehensive aerodynamic analysis. Larger volumes might be relevant for passenger comfort or electric battery installation, reflecting the practical considerations in car design. 

\begin{wrapfigure}{r}{0.5\textwidth} 
    \centering
    \includegraphics[width=0.5\textwidth]{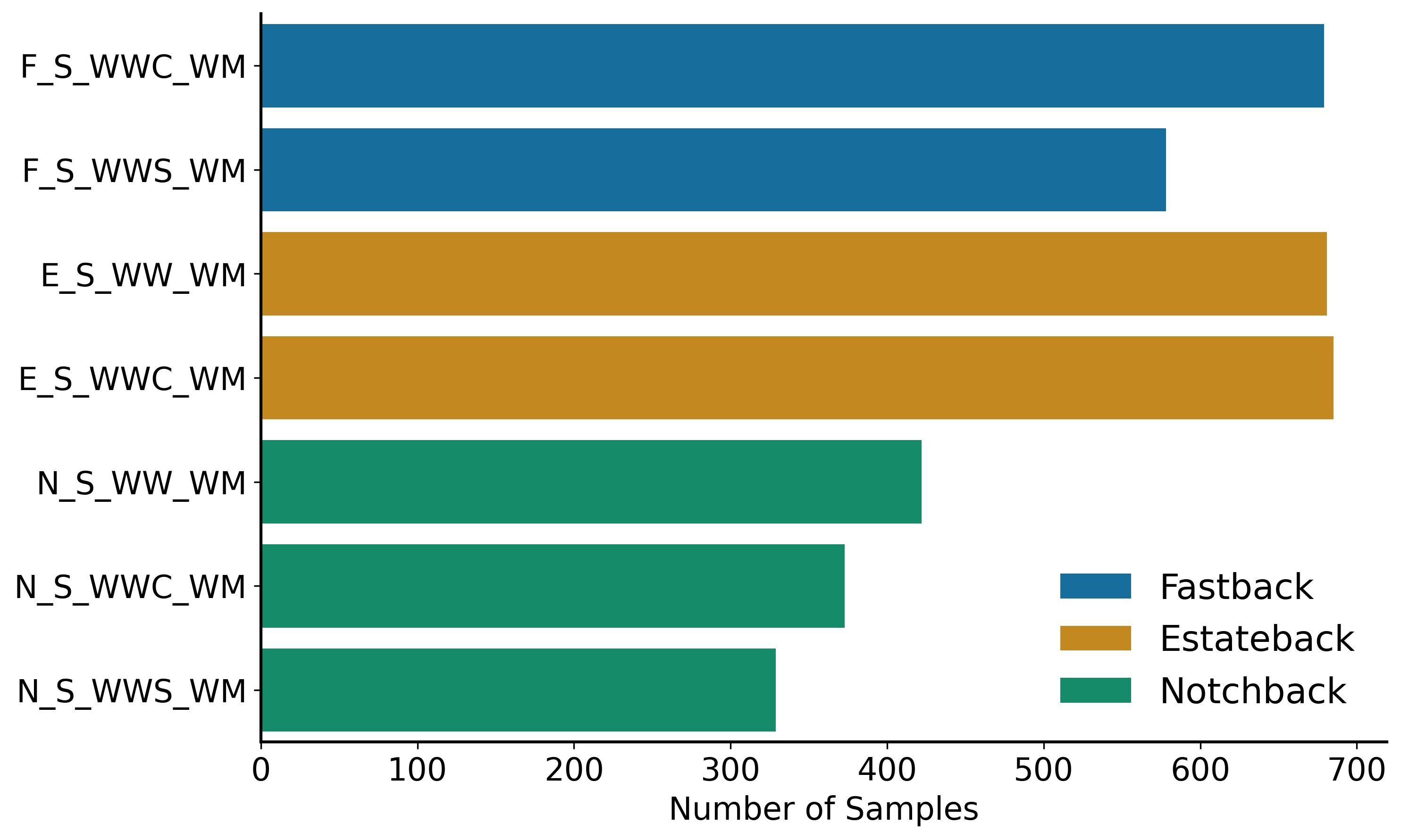}
    \caption{Sample counts for different car configurations in the DrivAerNet++ dataset. The figure shows the number of samples for each car category.}
    \label{fig:sample_counts}
    \vspace{-10pt}
\end{wrapfigure}
The parameters listed in Table \ref{tab:parameter_ranges} are used to morph the baseline geometries. Each parameter is categorized by group (A, B, C, etc.) and includes its minimum and maximum values, step sizes, range type, and units. Some design parameters are more relevant than others depending on the task; for example, A/B/C pillar thickness might be relevant for structural integrity and crashworthiness, while parameters like the diffusor angle are more critical for aerodynamics. Other parameters, such as windscreen inclination, are more relevant from an aesthetic perspective. This comprehensive parameterization ensures that the dataset captures diverse design attributes, enabling deep learning models to generalize well to out-of-distribution designs and flow field features.

\begin{table}[t]
\scriptsize
\centering
\setlength{\tabcolsep}{3pt} 
\caption{Design parameter names, step sizes, ranges, and units, categorized by group. These parameters are used for morphing the baseline geometries to create the DrivAerNet++ dataset. For Group I: Underbody (0 for smooth underbody, 1 for detailed underbody), Car Rear (0 for fastback, 1 for estateback, 2 for notchback), and Wheels (0 for open detailed, 1 for open smooth, 2 for closed smooth).}

\begin{tabular}{cclccccc}
\hline
\textbf{\#} & \textbf{Group} & \textbf{Parameter Name} & \textbf{Minimum Value} & \textbf{Maximum Value} & \textbf{Step Value} & \textbf{Range} & \textbf{Units} \\
\hline
1 & \multirow{4}{*}{A} & Car Width & -60 & 150 & & Bounds & mm  \\
2 &  & Car Length & -60 & 130 & & Bounds & mm \\
3 &  & Car Roof Height & -60 & 90 & & Bounds & mm \\
4 &  & Car Green House Angle & -200 & 150 & & Bounds & °\\ \hline
5 & \multirow{3}{*}{B} & Diffusor Angle & -8 & 15 & & Bounds & ° \\
6 &  & Ramp Angle & -8 & 15 & & Bounds & ° \\
7 &  & Trunklid Angle  & -8 & 20 & & Bounds & ° \\ \hline
8 & \multirow{3}{*}{C} & Side Mirrors Rotation & -20 & 20 & & Bounds & ° \\
9 &  & Side Mirrors Translate X & -10 & 20 & & Bounds & mm \\
10 &  & Side Mirrors Translate Z & -5 & 10 & & Bounds & mm \\ \hline
11 & \multirow{4}{*}{D} & Rear Window Inclination  & -1.8 & 2.5 & & Bounds & ° \\
12 &  & Rear Window Length  & -95 & 150 & & Bounds & mm \\
13 &  & Windscreen Inclination & -1.8 & 2.5 & & Bounds & ° \\
14 &  & Windscreen Length & -50 & 50 & & Bounds & mm \\ \hline
15 & \multirow{2}{*}{E} & A B C Pillar Thickness  & -15 & 5 & 0.1 & Step & \% \\
16 &  & Fenders Arch Offset & -25 & 55 & & Bounds & mm \\ \hline
17 & \multirow{3}{*}{F} & Door Handles Thickness & -20 & 30 & & Bounds & mm \\ 
18 &  & Door Handles X Position & -50 & 50 & & Bounds & mm \\
19 &  & Door Handles Z Position & -30 & 30 & & Bounds & mm \\ \hline
20 & \multirow{2}{*}{G} & Trunklid Curvature  & -0.4 & 0.8 & 0.1 & Step & \% \\
21 &  & Trunklid Length  & -40 & 40 & & Bounds & mm \\ \hline
22 & \multirow{2}{*}{H} & Front Bumper Curvature & -0.4 & 1 & & Bounds & \% \\
23 &  & Front Bumper Length & -25 & 60 & & Bounds & mm \\ \hline
24 & \multirow{3}{*}{I} & Underbody & 0 & 1 & & Binary & 0/1 \\ 
25 &  & Car Rear & 0 & 2 & & Options & 0/1/2 \\
26 &  & Wheels & 0 & 2 & & Options & 0/1/2 \\ \hline
\end{tabular}
\label{tab:parameter_ranges}
\end{table}

Figure~\ref{fig:sample_counts} presents the distribution of samples across different car categories within the DrivAerNet++ dataset. The first letter in the configuration name refers to the car body type: \textbf{F} for Fastback, \textbf{E} for Estateback, and \textbf{N} for Notchback. The second letter indicates the underbody type: \textbf{S} for Smooth underbody. The third letters describe the wheels configuration: \textbf{WWC} for with wheels closed, \textbf{WWS} for with wheels open smooth, and \textbf{WW} for with wheels open detailed. The fourth letters denote the presence of mirrors: \textbf{WM} for with mirrors. This detailed classification ensures precise categorization and analysis of the different car designs within the dataset. For comparison, the DrivAerNet~\cite{elrefaie2024drivaernet} was based only on the fastback category with detailed underbody, with open detailed wheels, and with mirrors \text{F\_D\_WW\_WM}.

\begin{figure}[h]
    \centering
    \includegraphics[width=\textwidth]{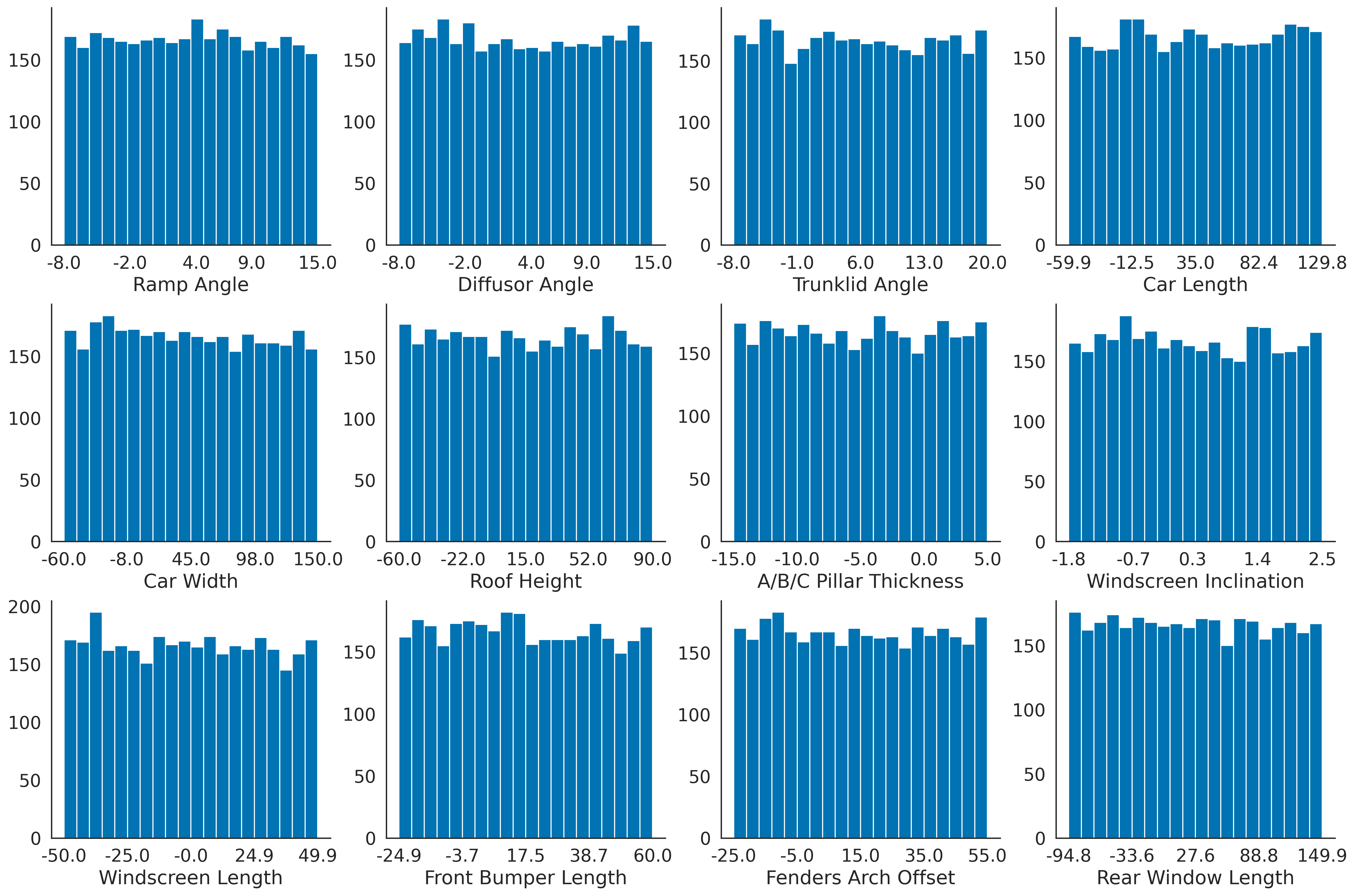}
    \caption{Distribution of various geometric parameters in the DrivAerNet++ dataset. The histograms demonstrate the range and frequency of these parameters, showcasing the diversity in our dataset.}
    \label{fig:parameter_distributions}
\end{figure}

In Figure \ref{fig:parameter_distributions}, we show the statistics of the design parameters in the DrivAerNet++ dataset. Each subplot represents a different design parameter.  

\vspace{-10pt}

\subsection{CFD Meshing}
\label{Appendix:Meshing}

The meshing procedure was conducted using the \texttt{snappyHexMesh} utility~\cite{openfoam2023snappyhexmesh}. This process involved three main stages: castellated mesh generation, snapping, and layer addition, ensuring a high-quality, boundary-fitted mesh suitable for accurate aerodynamic simulations.
Firstly, the geometry components, including the car body, front wheels, and rear wheels, were defined as triangular surface meshes.

\begin{figure}[h]
    \centering
    \includegraphics[width=\textwidth]{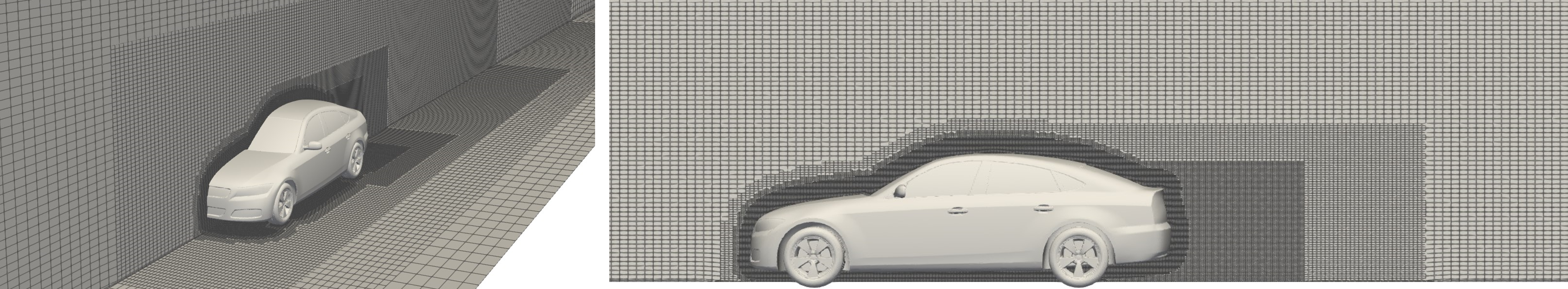}
  \caption{Mesh generation for the fastback model using SnappyHexMesh (SHM)~\cite{openfoam2023snappyhexmesh}. The left view shows a perspective of the car model within the computational domain, while the right view highlights the mesh refinement around the car body from a side perspective. The computational grid in the \( y \)-normal symmetry plane reveals four distinct regions of refinement, with additional layers surrounding the car body. These refinements are designed to capture both wake dynamics and boundary layer development accurately.}

    \label{fig:DrivAerNet_Meshing}
\end{figure}
In the castellated mesh generation stage, the mesh was initially refined around the surfaces, with the car body and wheels receiving high refinement levels (up to level 5). The snapping stage involved aligning the mesh with the geometry surfaces. Patch smoothing iterations ensured the mesh points conformed accurately to the surface features. Five mesh relaxation iterations were performed to achieve the correct mesh alignment. The final stage, layer addition, involved growing prism layers on the car body, wheels, and ground surfaces. Five layers of different resolutions were applied adjacent to the car body, with the rear region receiving additional refinement to accurately capture the detaching and reattaching flow.

\begin{figure}[h]
    \centering
    \includegraphics[width=\textwidth]{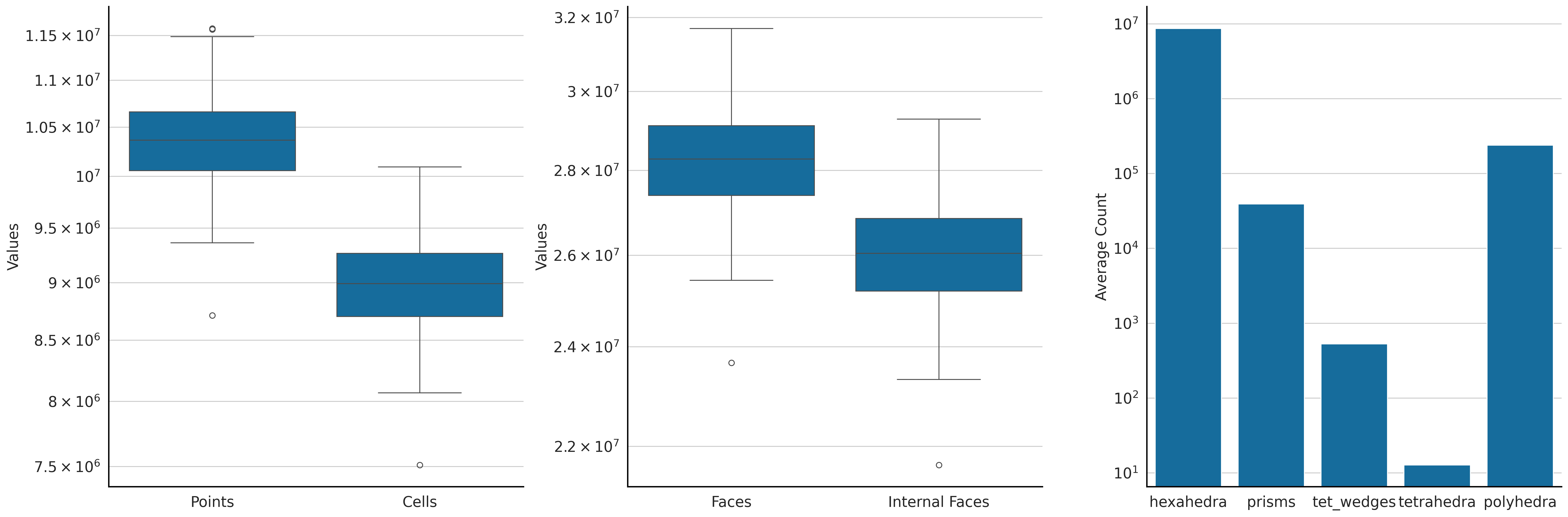}
    \caption{Mesh statistics for the DrivAerNet++ dataset. Left: distribution of points and cells; middle: distribution of faces and internal faces; right: average cell counts by type.}
    \label{fig:mesh_statistics}
\end{figure}

Volumes of refinement (VoR) were specified using searchable boxes with precise coordinates to target critical areas for mesh refinement, as shown in Figure \ref{fig:DrivAerNet_Meshing}. Instead of using one mesh template for all simulations, we developed a mesh refinement strategy based on the car shape, category, and dimensions. The size, dimensions, and position of the refinement boxes were adjusted according to the car geometry and category to ensure accurate modeling of the wake flow. 
Figure \ref{fig:mesh_statistics} illustrates the variation in mesh sizes within the DrivAerNet++ dataset. The meshes in DrivAerNet++ consist mainly of hexahedra cells, along with prisms, tetrahedral wedges, tetrahedra, and polyhedra. The distribution of points, cells, faces, and internal faces is also illustrated, providing an overview of the mesh quality and complexity.

\subsection{Numerical Simulation}

\subsubsection{Domain and Boundary Conditions}
For the CFD simulations, we used the open-source software OpenFOAM® v11,\footnote{\url{https://openfoam.org/version/11/}} a comprehensive suite of C++ modules that supports solver customization, utility development, and parallelization in CFD studies. In the field of aerodynamics, only two open-source tools—OpenFOAM~\cite{OpenFOAMv11} and SU2~\cite{su2}—are commonly used, with OpenFOAM being the state-of-the-art tool most widely adopted in both industry and academia for car aerodynamics. A review of the literature indicates no significant technical advantages of using SU2 over OpenFOAM for this application. We utilized 1:1 scaled models of all car designs and the SIMPLE algorithm (Semi-Implicit Method for Pressure Linked Equations) to couple pressure and velocity in the incompressible fluid simulations. These were carried out using the \textit{simpleFoam} solver, which is ideal for steady-state, turbulent, and incompressible flow simulations (In OpenFOAM® v11, the \textit{incompressibleFluid} modular solver replicates \textit{simpleFoam}).

We selected the state-of-the-art $k$-$\omega$-SST turbulence model based on Menter's formulation~\cite{menter2003ten} for the Reynolds-Averaged Navier-Stokes (RANS) simulations. This model is widely used in both academic research and industry due to its ability to overcome the limitations of the standard $k$-$\omega$ model, particularly its sensitivity to the freestream values of $k$ and $\omega$, and its effectiveness in predicting flow separation.

Following a similar approach to \cite{heft2012experimental}, we set the solver tolerances to $10^{-6}$ for pressure ($p$), $10^{-8}$ for velocity ($U$) and turbulence quantities ($k$ and $\omega$), and $10^{-7}$ for the potential solver. The relative solver tolerance, defined as the ratio between the current and initial residuals, was set to $10^{-2}$ for pressure, $10^{-3}$ for velocity and turbulence quantities, and $10^{-2}$ for the potential solver.
The simulations were conducted at a flow velocity (\(u_{\infty}\)) of 30 m/s (108 km/h), corresponding to a Reynolds number range of approximately \(8.366 \times 10^6\) to \(1.006 \times 10^7\), using the car's length as the characteristic length scale. These Reynolds numbers are based on the car with the smallest length (4344 mm) and the largest length (5210 mm). 
  All simulations were performed using 12 million cells (refer to the meshing section for more details), and due to the symmetry, this amounts to a total of 24 million cells for the full simulation. Additional layers were added around the car body to accurately capture wake dynamics and boundary layer evolution (see Figure \ref{fig:DrivAerNet_Meshing}). Different meshing and refinement strategies were employed based on the car category, as the rear end varies, necessitating distinct meshing at the back of the car to account for flow separation and accurately model the wake flow. For the car underbody, we differentiated between smooth and detailed underbody configurations.

\begin{figure}[h!]
    \centering
    \includegraphics[width=0.7\textwidth]{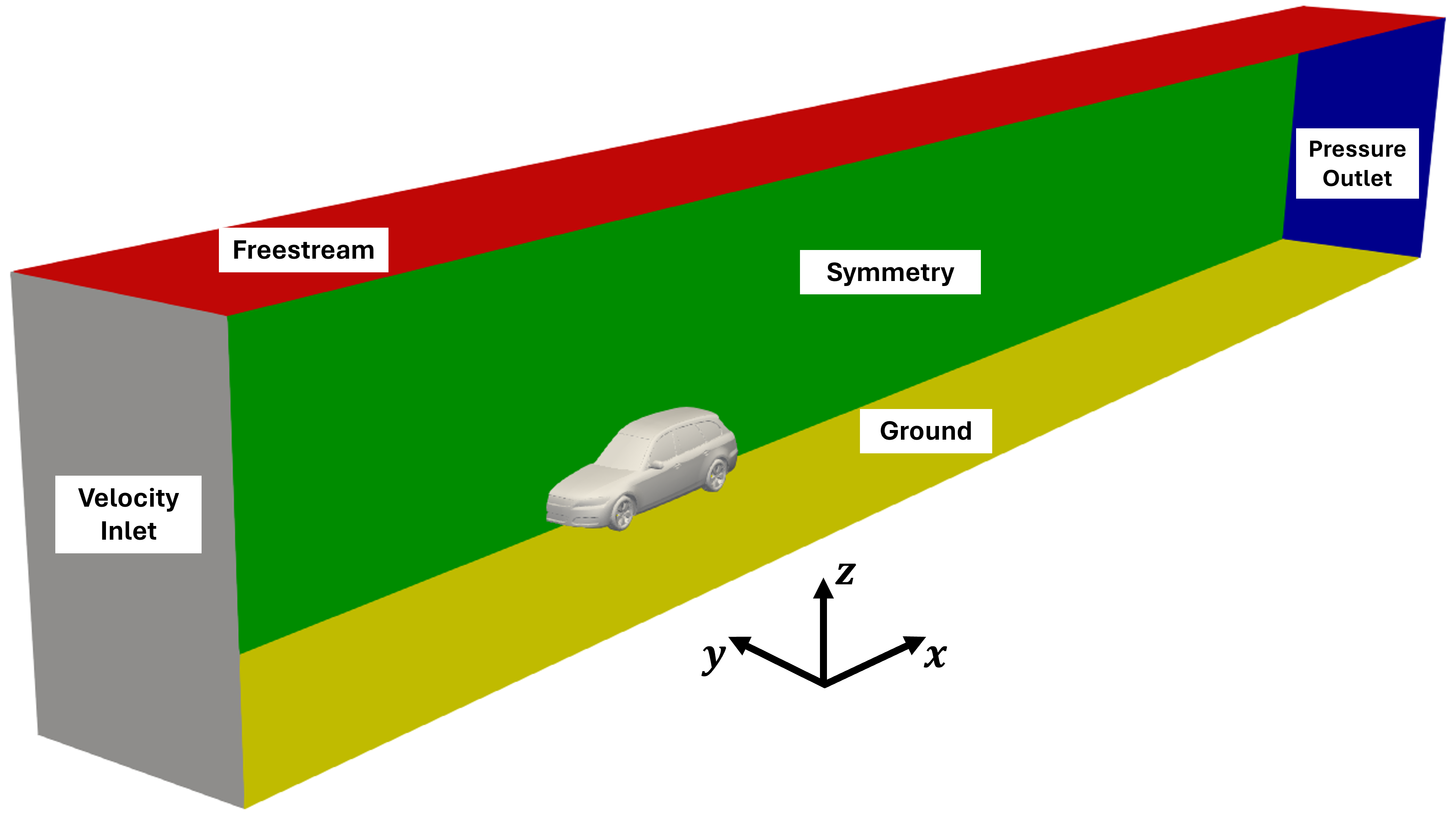}
\caption[CFD simulation domain and boundary conditions]{CFD simulation domain and boundary conditions used for our DrivAerNet++ dataset. The domain includes the velocity inlet (grey), pressure outlet (blue), freestream (red), symmetry (green), ground (yellow), and side wall (not shown) boundaries. The car surface was assigned no-slip conditions, the ground was assigned a moving wall condition, and the wheels were modeled with a rotating wall velocity boundary condition.}
    \label{fig:CFD_Domain}
\end{figure}

Boundary conditions were defined with a uniform velocity at the inlet and pressure-based conditions at the outlet. To prevent backflow into the simulation domain, the velocity boundary condition at the outlet was set as an \textit{inletOutlet} condition. The car surface was assigned no-slip conditions, and the ground was assigned a moving wall condition, while the wheels were modeled to rotate with the \textit{rotatingWallVelocity} condition, defined as:
\begin{equation}
u_{\text{wheels}} = \omega \times r
\end{equation}
where $\omega$ is the angular velocity and $r$ is the wheel radius. Slip conditions were applied to the lateral and top boundaries of the domain. 

As illustrated in Figure~\ref{fig:CFD_Domain}, the domain includes the velocity inlet, pressure outlet, freestream, symmetry, ground, and side wall boundaries. Compared to DrivAerNet~\cite{elrefaie2024drivaernet}, we utilized the symmetry boundary condition, which leverages the symmetry of the car and domain. This allowed us to double the mesh resolution while maintaining the same computational time for the symmetric car designs. 

Near-wall viscosity effects were addressed using the \textit{nutUSpaldingWallFunction} wall function approach. This wall function for the viscosity term applies a continuous turbulent viscosity profile near the wall, based on velocity, as proposed by~\cite{spalding1974numerical}. For divergence terms, the default Gauss linear scheme was utilized, with the velocity convective term discretized using a bounded Gauss \textit{linearUpwindV} scheme to ensure second-order accuracy. Gradient calculations were performed using the Gauss linear method, complemented by a multi-dimensional limiter to enhance solution stability. The primary quantities of interest were the 3D flow field, surface pressure, wall-shear stresses, 3D streamlines, and aerodynamic coefficients.

\begin{table}[h]
\scriptsize
\centering
\caption{Definition of the quantities and their values for air at sea level and at a temperature of 298.15 K (25 $^\circ$C).}
\begin{tabular}{@{}lll@{}}
\toprule
\textbf{Quantity} & \textbf{Definition}                              & \textbf{Value}            \\ \midrule
$\rho$            & Density of air                           & 1.184 kg m$^{-3}$         \\
$\nu$             & Kinematic viscosity of the fluid                 & $1.56 \times 10^{-5}$ m$^2$ s$^{-1}$ \\
$u_{\infty}$        & Inlet velocity                            & 30 m/s (108 km/h)                         \\
$Re_L$            & Reynolds number               & \(8.366-10.06 \times 10^6\)       \\
 \bottomrule
\end{tabular}
\label{tab:PhysicalQuantities}
\end{table}

We used the \textit{potentialFoam} solver to initialize the OpenFOAM®  simulations. This solver calculates a potential flow solution to quickly establish an initial velocity field, thereby accelerating the convergence of the solution in complex aerodynamic simulations~\cite{OpenFOAMv11}. The potential flow assumption neglects viscous effects and assumes irrotational flow, providing a good starting point for the \textit{simpleFoam} solver.
The simulations were conducted using air with the properties listed in Table~\ref{tab:PhysicalQuantities}. 

\subsubsection{Meshing Sensitivity}
Mesh sensitivity analysis is conducted to ensure that simulation results are independent of the mesh resolution, which can significantly impact the accuracy of the CFD simulations. By varying the mesh density and observing its effect on outputs like the drag coefficient, analysts can determine the optimal mesh size that balances computational cost with solution accuracy. This process helps to confirm that the results are not artifacts of the chosen mesh and improves confidence in the reliability of the simulation predictions.
We conducted a mesh sensitivity analysis to examine the effect of mesh resolution on the drag coefficient specifically for the fastback configuration. The findings are summarized in Table~\ref{tab:mesh_resolution}, which shows that as the mesh quality improves from coarse (6M cells) to fine (24M cells), the relative error in drag coefficient decreases from 8.21\% to 2.17\%, indicating a convergence towards the reference CFD results from TUM~\cite{heft2012experimental}. However, the computational time increases significantly with finer meshes, demonstrating the trade-off between accuracy and computational cost. This analysis highlights the importance of selecting an appropriate mesh resolution to balance precision with efficiency in CFD simulations.
\begin{table}[h]
\scriptsize
\centering
\caption{Influence of mesh resolution on drag coefficient for the fastback case. The relative error is computed with respect to the CFD results from TUM~\cite{heft2012experimental}.}
\vspace{3pt}
\begin{tabular}{lccc}
\hline
Mesh Quality & Number of Cells & Relative Time & Relative Error in Drag (\%) \\ \hline
Coarse  & 6M   & 1.00 & 8.21\% \\ 
Medium  & 12M  & 1.25 & 6.55\% \\ 
Fine    & 24M  & 1.88 & 2.17\% \\ \hline
\label{tab:mesh_resolution}
\end{tabular}
\end{table}

\subsubsection{Validation of the Numerical Results}

In this section, we present validation results for the baseline simulations. The assumption is that given a baseline geometry, conducting the CFD simulation and benchmarking the results against both reference simulations and wind tunnel experimental data should establish a foundation for using the same CFD meshing approach, boundary conditions, and solvers for the morphed geometries of this baseline geometry. However, DrivAerNet++ encompasses various car designs and categories, making a single mesh unsuitable due to variations in car length, rear end, and underbody features. Therefore, as detailed in the meshing section~\ref{Appendix:Meshing}, we employ different meshing strategies based on car categories. Similarly, we benchmark the CFD simulation results across different geometries, focusing on the following car categories:

\begin{enumerate}
    \item Fastback with open detailed wheels, detailed underbody, and mirrors
    \item Fastback with open detailed wheels, smooth underbody, and mirrors
    \item Notchback with open detailed wheels, smooth underbody, and mirrors
        \item Estateback with open detailed wheels, smooth underbody, and mirrors
\end{enumerate}

Based on this, the only factors that need adjustment for other car categories should be the geometry and meshing strategy, not the CFD solver or discretization scheme. To ensure convergence, 7000 iterations were performed, with forces averaged over the last 1000 iterations. We provide both the mean and standard deviation for uncertainty quantification in both the CFD and deep learning model predictions.

We compare our results with the original paper by the creators of the DrivAer body~\cite{heft2012experimental, heft2012introduction}. The primary goal is to determine whether the CFD modeling accurately predicts the correct trend of drag values within acceptable tolerance. The drag coefficient is defined as:

\begin{equation}
C_d = \frac{F_d}{\frac{1}{2} \rho u_{\infty}^2 A_{\text{ref}}}
\end{equation}

where $F_d$ represents the drag force experienced by the body, $A_{\text{ref}}$ is the effective frontal area, $u_{\infty}$ is the freestream velocity, and $\rho$ is the air density.

Table~\ref{tab:drag_coefficients} presents the comparison of drag coefficients ($C_D$) from our simulations, the reference CFD simulation, and wind tunnel experimental data provided by TUM. The simulation results show good agreement with the reference simulations and the experimental results, demonstrating a strong correlation between the drag values and ensuring a balance between simulation accuracy (fidelity) and computational cost (simulation time and mesh size).  For the fastback car with a smooth underbody, the difference is less than 3$\%$, and with the detailed body, we have a difference of 2.18$\%$. 
For both the notchback and estateback configurations, the differences are still below 5$\%$. The estateback configuration exhibits the highest drag coefficient, whereas the difference in drag coefficients between the fastback and notchback configurations is minimal.

\begin{table}[h]
\scriptsize
\centering
\caption{Comparison of drag coefficients ($C_D$) from experiments and CFD simulations.}
\begin{tabular}{lccccc}
\hline
 & Experiment TUM & Reference CFD TUM & Ours & Difference Exp. & Difference CFD \\
\hline

Fastback smooth & 0.243 & 0.241 & 0.236 & 2.88\% & 2.07\% \\
Fastback detailed & 0.275 & 0.278 & 0.269 & 2.18\% & 3.24\% \\
Notchback & 0.246 & n/a &0.234 & 4.88\% & n/a \\
Estateback & 0.292 & n/a & 0.280 & 4.11\% & n/a \\
\hline
\end{tabular}
\label{tab:drag_coefficients}
\end{table}

The results in Table~\ref{tab:drag_coefficients_delta} demonstrate our model's ability to accurately predict differences in drag coefficients ($\Delta C_D$) between various car configurations. For instance, the predicted drag difference between the fastback detailed (FD) and fastback smooth (FS) configurations is 0.033, closely aligning with the experimental difference of 0.032. Similarly, the difference between the estateback (E) and notchback (N) configurations is exactly predicted as 0.046, matching the experimental result. However, some discrepancies, such as the difference between the estateback and fastback smooth configurations (0.039 vs. 0.049 experimentally), indicate areas for further refinement. Overall, the model shows strong alignment with experimental data for most configurations, with deviations generally within an acceptable range, highlighting the balance between simulation accuracy and the computational cost required.

\begin{table}[h]
\scriptsize
\centering
\caption{Comparison of differences in drag coefficients ($\Delta C_D$) from experiments and CFD simulations.}
\begin{tabular}{lccc}
\hline
 & Experiment TUM & Reference CFD TUM & Ours  \\
\hline
FD - FS & 0.032 & 0.037 & 0.033  \\
FD - N  & 0.029 & n/a   & 0.035  \\
E - FD  & 0.017 &  n/a  & 0.011 \\
N - FS  & 0.003 &  n/a  & $-$0.002 \\
E - FS  & 0.049 & n/a & 0.039  \\
E - N  & 0.046 &  n/a  &  0.046 \\
\hline
\end{tabular}
\label{tab:drag_coefficients_delta}
\end{table}

\subsection{Convergence Detection}

For detecting the convergence of each CFD case, we implemented the following approaches:

\begin{itemize}
\item \textbf{Geometry Quality}: Ensuring car designs are watertight and free from intersecting faces or holes is critical. We use the functionalities provided by ANSA® and Blender~\cite{blender} software for geometry cleaning and processing. In addition, we ensure that the 3D meshes are correctly aligned within the simulation domain and that the wheelbase distance between the wheels is accurately defined, which is necessary for the \textit{rotatingWallVelocity} boundary condition. Additionally, we ensure the correct estimation of the car's frontal projected area, which is essential for accurate drag coefficient calculations.

\item \textbf{CFD Mesh Quality}: For mesh quality, we employ the \textit{checkMesh} utility from OpenFOAM®. This utility verifies various aspects such as the overall domain bounding box, geometric and solution directions, boundary and cell openness, aspect ratio, face and cell volumes, mesh non-orthogonality, face pyramids, skewness, and coupled point location match. These checks ensure that the mesh meets predefined criteria, guaranteeing high quality and suitability for accurate CFD simulations, as detailed in the OpenFOAM® documentation.

    \item \textbf{Drag Convergence Detection}: Using a runtime control algorithm, we average drag values and provide statistical errors (mean and standard deviation) to assist in uncertainty quantification and improve deep learning model training.
    \item \textbf{Residual Monitoring}: Monitoring residuals for main physical quantities, ensuring convergence when all RMS residual values are below $10^{-5}$, which indicates a well-converged numerical solution.
    \item \textbf{Outlier Detection}: Identifying and removing outlier designs that passed quality checks but adversely affected model predictions during training and validation.
\end{itemize}

\section{In-Depth Insights from DrivAerNet++}

\begin{figure}[h]
    \centering
    \includegraphics[width=\textwidth]{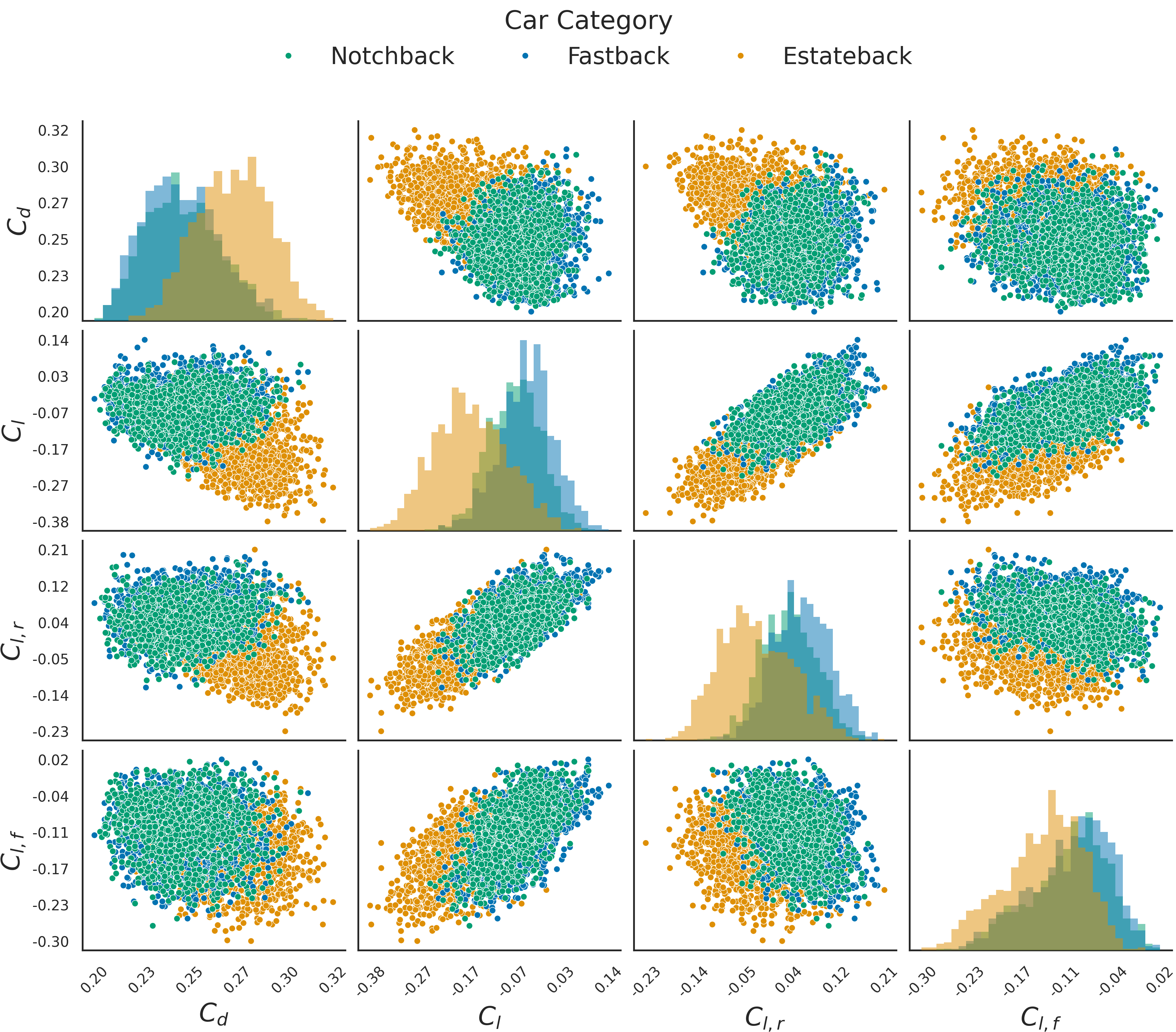}
        \caption{Pair plot showing the relationships between different aerodynamic coefficients (\(C_d\), \(C_l\), \(C_{l,r}\), and \(C_{l,f}\)) in the DrivAerNet++ dataset. The diagonal histograms represent the distribution of each coefficient, while the scatter plots show the pairwise correlations between the coefficients. The colors represent different configurations within the dataset, illustrating the diversity in aerodynamic performance.}
\label{fig:pairplots_DrivAerNet}
\end{figure}

\begin{figure}[h!]
    \centering
    \includegraphics[width=\textwidth, height=0.95\textheight]{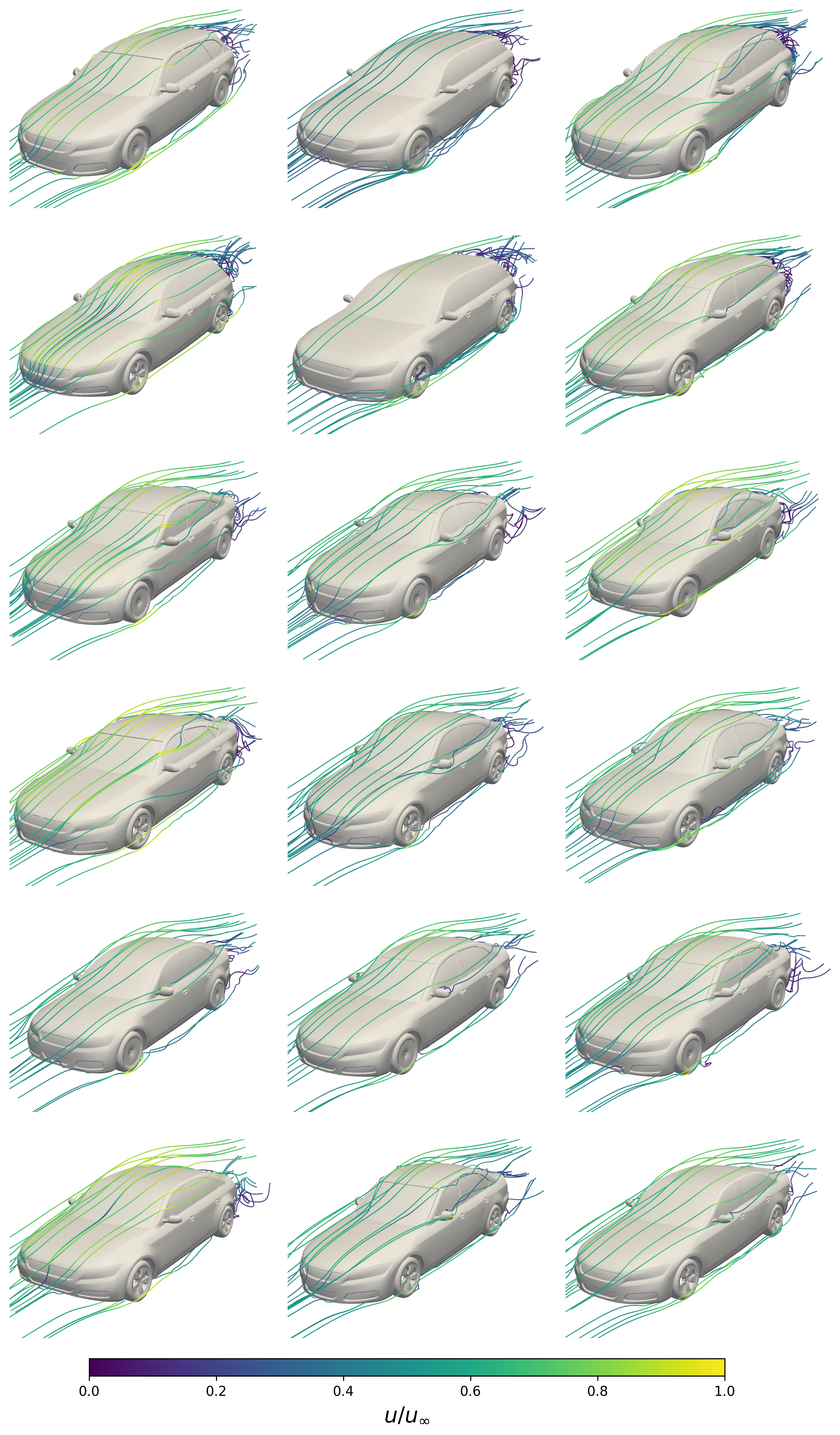}
    \caption{Streamlines visualizations for different car designs, showcasing the large variations in shape within the DrivAerNet++ dataset. The color indicates the normalized velocity \(u/u_{\infty}\), with values ranging from 0 to 1.}
    \label{fig:DrivAerNet_streamlines}
\end{figure}

\begin{figure}[h!]
    \centering
    \includegraphics[width=\textwidth, height=0.95\textheight]{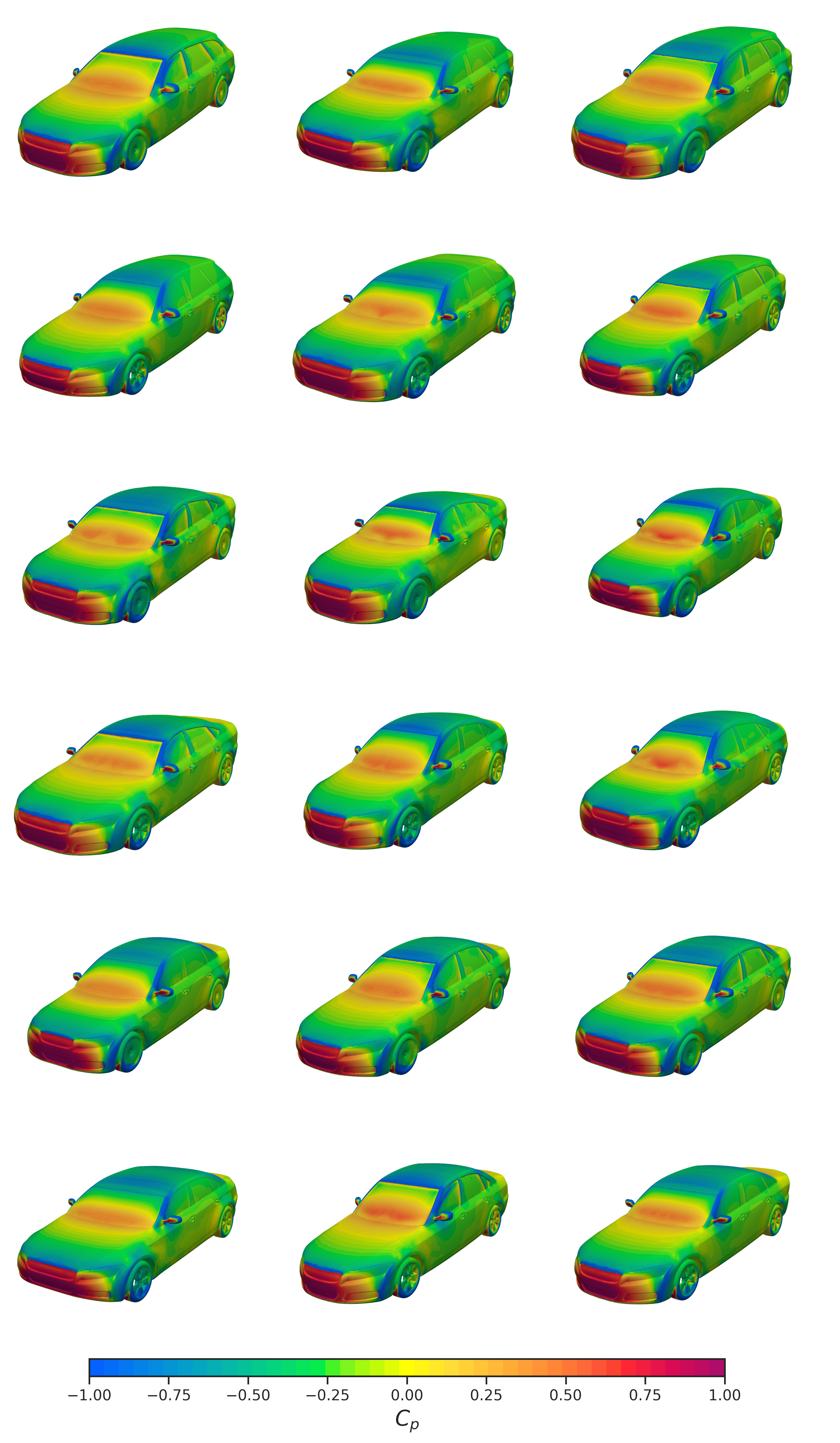}
    \caption{Pressure coefficient (\(C_p\)) visualizations for different car designs, illustrating the large variations in shape within the DrivAerNet++ dataset. The color indicates the pressure coefficient \(C_p\), with values ranging from -1.00 to 1.00.}
    \label{fig:DrivAerNet_cp}
\end{figure}

\begin{figure}[h!]
    \centering
    \includegraphics[width=\textwidth, height=0.95\textheight]{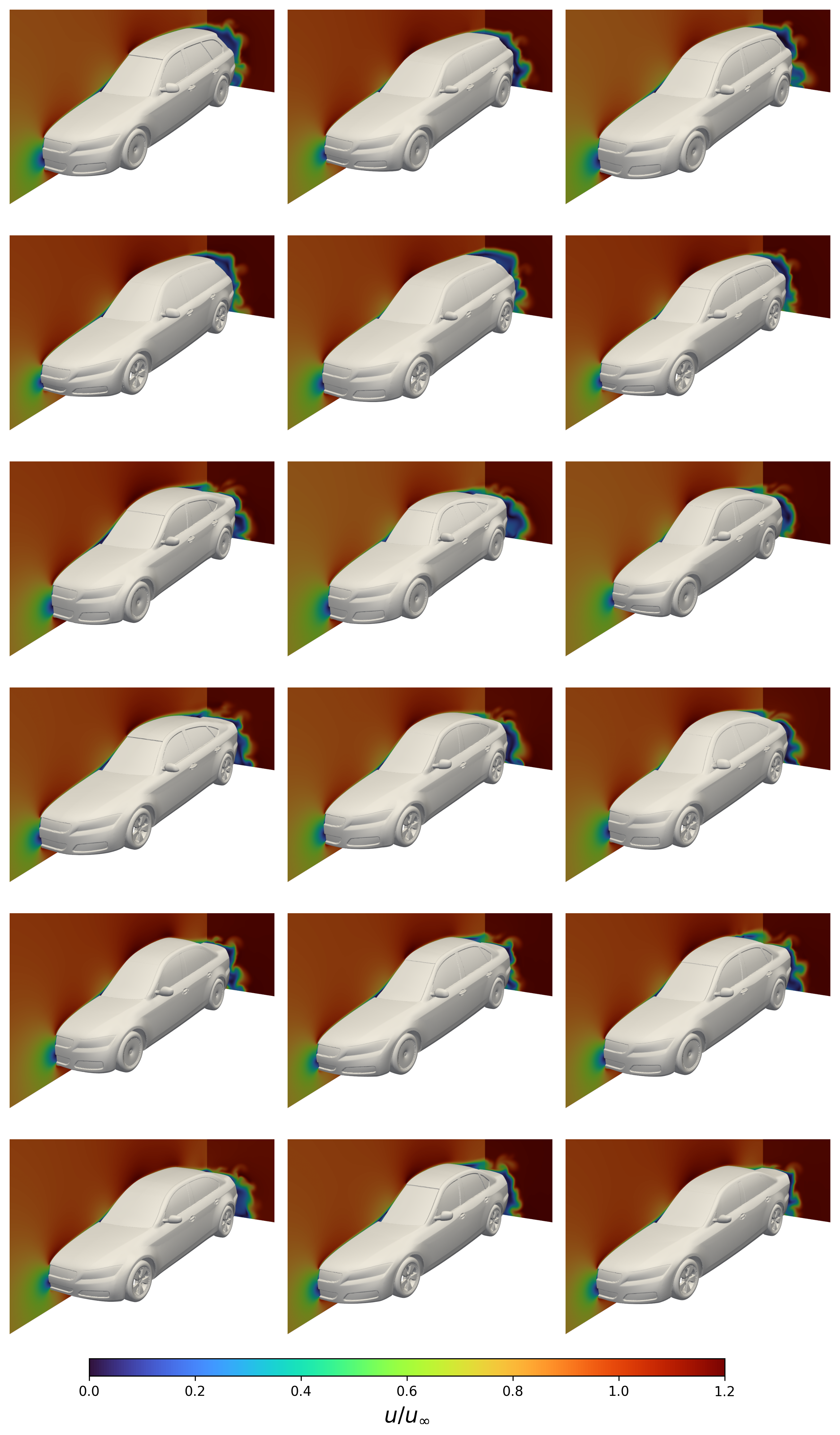}
    \caption{Visualizations of the 3D flow field around different car designs, showcasing the large variations in shape within the DrivAerNet++ dataset. The color indicates the normalized velocity \(u/u_{\infty}\), with values ranging from 0 to 1.2.}
    \label{fig:DrivAerNet_flowfield}
\end{figure}

In this section, we provide an overview of the contents and results from the CFD simulations included in our DrivAerNet++ dataset, highlighting the diversity and richness of the data. This includes pair plots of aerodynamic coefficients (Figure~\ref{fig:pairplots_DrivAerNet}), visualizations of streamlines (Figure~\ref{fig:DrivAerNet_streamlines}), pressure distribution on the car surface (Figure~\ref{fig:DrivAerNet_cp}), and the 3D velocity field around the car (Figure~\ref{fig:DrivAerNet_flowfield}), illustrating the comprehensive range of aerodynamic behaviors captured in the dataset.

\clearpage

\clearpage

\subsection{Dataset Annotations}
In addition to the CFD simulation data, our dataset includes detailed annotations for various car components, such as wheels, side mirrors, and doors. These annotations are instrumental for a range of machine learning tasks, including classification, semantic segmentation, and object detection. The comprehensive labeling can also facilitate automated CFD meshing processes by providing precise information about different car components. By incorporating these labels, our dataset enhances the utility for developing and testing advanced algorithms in automotive design and analysis.

\begin{figure}[h!]
    \centering
    \includegraphics[width=\textwidth]{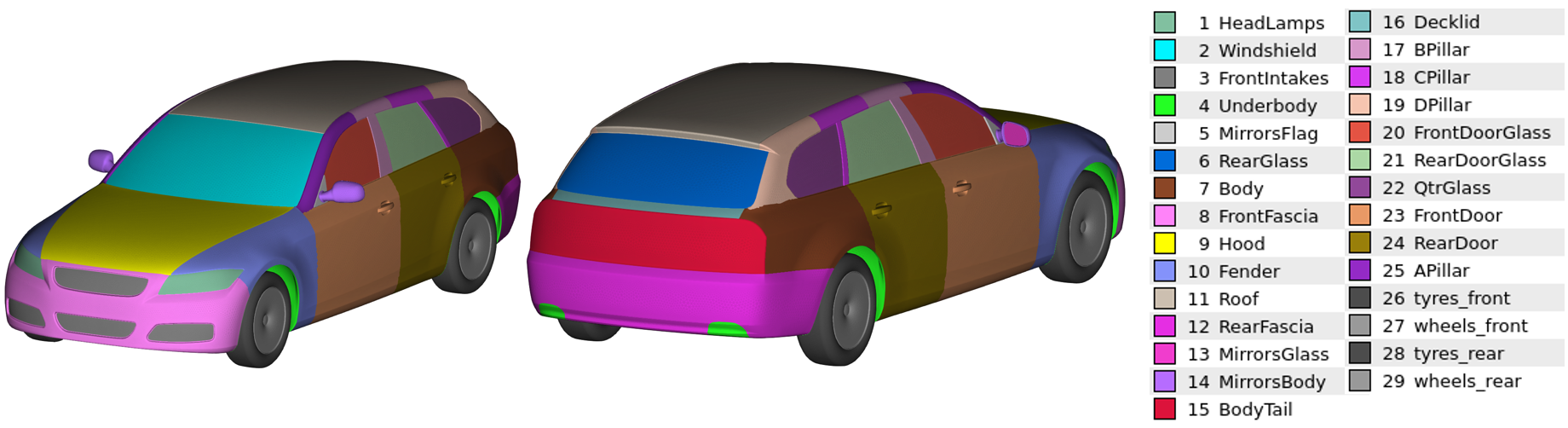}
    \caption{Annotated car components in the dataset, featuring 29 labels. Each color represents a different part of the car, including headlamps, windshield, front intakes, underbody, mirrors, rear glass, body panels, and wheels, providing a comprehensive labeling of the full car geometry.}
    \label{fig:annotated_car_components}
\end{figure}

As illustrated in Figure~\ref{fig:annotated_car_components}, the dataset comprises annotations for 29 distinct car components, with each color representing a different part.  This extensive annotation enables detailed analysis and optimization of aerodynamic performance, structural integrity, and aesthetic design, thus supporting a wide range of engineering and research applications.

\section{Deep Surrogate Models}

\subsection{RegDGCNN}

In our previous work~\cite{elrefaie2024drivaernet}, we introduced the RegDGCNN model for an end-to-end aerodynamic drag prediction directly from 3D meshes.  Our RegDGCNN model combines PointNet's spatial encoding~\cite{qi2017pointnet} with graph CNNs' relational analysis to understand fluid dynamics. It uses edge convolution (EdgeConv) on dynamically updating graphs to capture complex fluid interactions, providing a novel approach for precise aerodynamic parameter estimation.

The RegDGCNN model is designed for regression tasks involving 3D point cloud data, leveraging graph-based convolutions to extract hierarchical features effectively~\cite{wang2019dynamic}. The network architecture comprises several key components: EdgeConv layers utilize local neighborhood information to perform convolution operations on graph-structured data, with four EdgeConv layers featuring increasing channel sizes of 256, 512, 512, and 1024. Each layer is followed by batch normalization~\cite{ioffe2015batch} and Leaky ReLU activation to stabilize and accelerate training.  Dropout is applied in the fully connected layers to prevent overfitting, with a dropout rate of 0.4. After extracting features using EdgeConv layers, the network includes fully connected layers that progressively reduce the dimensionality of the feature vectors, with sizes of 128, 64, 32, and 16 neurons, respectively. The final output layer predicts the drag coefficient. Additionally, the network employs both global max pooling and global average pooling to aggregate features from all points in the point cloud, ensuring that the model captures the most critical information.

\subsection{PointNet}

We modified the original PointNet~\cite{qi2017pointnet} network for the task of aerodynamic drag prediction. The modified model consists of a structured sequence of layers, beginning with an input layer for 3D coordinates, followed by six convolutional layers with increasing channel sizes: 64, 128, 256, 256, 512, and up to the embedding dimension, which we set to 1024. Each convolutional stage includes batch normalization~\cite{ioffe2015batch} for enhanced training stability. A residual connection links the input directly to the deeper layers, aiding gradient flow. The architecture concludes with three linear layers, reducing dimensions progressively from 1024 to 512, and finally to a single output value for the drag.

Compared to RegDGCNN~\cite{elrefaie2024drivaernet}, we scaled the number of point clouds up to 100k and 250k, which showed an increase in performance compared to the baseline implementation, typically using 1-5k points. The increased number of points provides a good representation of the actual geometry and can better capture small geometric modifications. The network is fully differentiable and can be integrated with downstream optimization tasks.

We also considered another variant of PointNet for the task of pressure prediction on the car surface (see Section~\ref{subsec:pressure_prediction}).  The PointNet model for pressure prediction consists of an input layer for 3D coordinates, followed by three convolutional layers with increasing channel sizes: 64, 128, and 1024.  A Spatial Transformer Network (STN)~\cite{jaderberg2015spatial} is used to align the input points, enhancing the network's ability to learn spatial features. 

\subsection{Graph Neural Network}
Here, we train a graph neural network (GNN) for predicting the aerodynamic drag. This model is implemented using the PyTorch Geometric library~\cite{fey2019fast} and consists of several layers and components tailored for processing graph-structured data. The architecture of this model comprises several key components. The model begins with a series of four Graph Convolutional Network (GCN) layers, which use the \textit{GCNConv} operator~\cite{kipf2016semi} for message passing to capture local structural information from the graph. These layers have increasing channel sizes of 64, 128, 128, and 256, respectively. Following the GCN layers, the model includes a fully connected network with three linear layers, which progressively reduce the feature dimensions to 128, 64, and finally to a single output value for drag prediction. ReLU activations and batch normalization~\cite{ioffe2015batch} are used between the layers. To aggregate the node features into a graph-level representation, global mean pooling is applied, computing the mean of the node features for each graph in the batch.

The training was distributed across four NVIDIA A100 80GB GPUs, leveraging data parallelism for increased computational efficiency. The initial learning rate was set to 0.001, with a learning rate scheduler employed to reduce the rate when validation loss plateaued. Specifically, the \textit{ReduceLROnPlateau} scheduler was used, with a patience of 10 epochs and a reduction factor of 0.1. This strategy helped fine-tune the models by adjusting the learning rate based on validation performance. The training process spanned 100 epochs. The Adam optimizer~\cite{kingma2014adam} was chosen for its adaptive learning rate capabilities.

\section{Evaluation}

We use various loss functions to assess model performance in predicting both the aerodynamic drag and the pressure values on the car's surface. Below are the key metrics:
\newline
\textbf{Mean Squared Error (MSE):}
This metric quantifies the average of the squares of the errors, which are the differences between the ground truth drag coefficients from CFD simulations and the predictions made by the surrogate models. It is especially sensitive to large errors in prediction.
\begin{equation}
    MSE = \frac{1}{n} \sum_{i=1}^{n} (C_{d_{i}} - \hat{C}_{d_{i}})^2
\end{equation}
\textbf{Mean Absolute Error (MAE):}
The MAE measures the average magnitude of the errors in a set of predictions, without considering their direction. It is less sensitive to outliers than MSE.
\begin{equation}
    MAE = \frac{1}{n} \sum_{i=1}^{n} |C_{d_{i}} - \hat{C}_{d_{i}}|
\end{equation}
\textbf{Maximum Absolute Error (Max AE):}
The Max AE measures the largest absolute error among all prediction errors, highlighting the worst-case prediction accuracy.
\begin{equation}
    Max\, AE = \max_{i=1}^{n} |C_{d_{i}} - \hat{C}_{d_{i}}|
\end{equation}
\textbf{Coefficient of Determination (\(R^2\) Score):}
The \(R^2\) Score indicates the proportion of variance in the actual drag coefficients that is predictable from the model's predictions, with a value of 1 representing a perfect fit.
\begin{equation}
    R^2 = 1 - \frac{\sum_{i=1}^{n} (C_{d_{i}} - \hat{C}_{d_{i}})^2}{\sum_{i=1}^{n} (C_{d_{i}} - \bar{C}_{d})^2}
\end{equation}

Lower MSE and MAE values, along with a higher \(R^2\) score and a lower Max MAE, indicate more accurate predictions of the aerodynamic drag coefficient by the surrogate models.

\section{Additional Experiments}

In this section, we utilize our dataset for various tasks to demonstrate its broad applicability. We first perform a full 3D surface field prediction of the pressure field. Then, we explore and visualize the design space of DrivAerNet++ using dimensionality reduction techniques such as t-SNE~\cite{van2008visualizing}.

\subsection{3D Surface Field Prediction}
\label{subsec:pressure_prediction}

\begin{wrapfigure}{r}{0.5\textwidth} 
    \centering
    \includegraphics[width=0.5\textwidth]{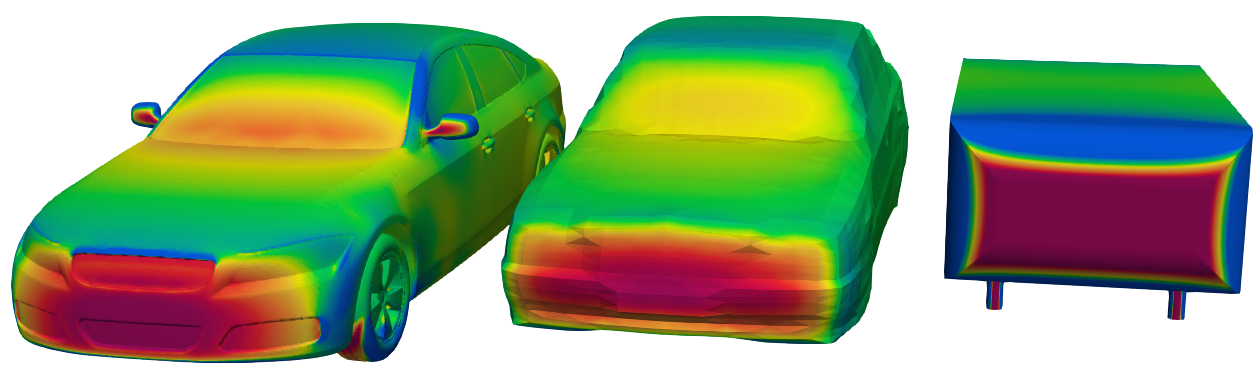}
    \caption{Pressure distribution on the car surface from three different datasets. The left car is from our dataset, DrivAerNet++, featuring realistic car shapes, higher mesh resolution, and greater simulation fidelity. The middle car is based on ShapeNet cars~\cite{Umetani2018}, and the right car uses the Ahmed body as presented in the study by Li et al. (2023)~\cite{li2023geometryinformed}.}
    \label{fig:Pcomparison}
    \vspace{-10pt}
\end{wrapfigure}

Predicting the pressure distribution on the 3D mesh surface of a car is an essential task for aerodynamic analysis. Detailed pressure distribution data around a car's surface allows aerodynamics experts to design modifications that reduce drag, optimize lift and downforce, prevent flow separation, enhance cooling efficiency, and minimize noise~\cite{hucho2013aerodynamik}. Here, we employ geometric deep learning models for surface field prediction based on our dataset. As illustrated in Figure~\ref{fig:Pcomparison}, our dataset, DrivAerNet++, provides a higher mesh resolution and more realistic car shapes compared to other datasets, with higher simulation fidelity.

\begin{figure}[h!]
    \centering
    \includegraphics[width=\textwidth]{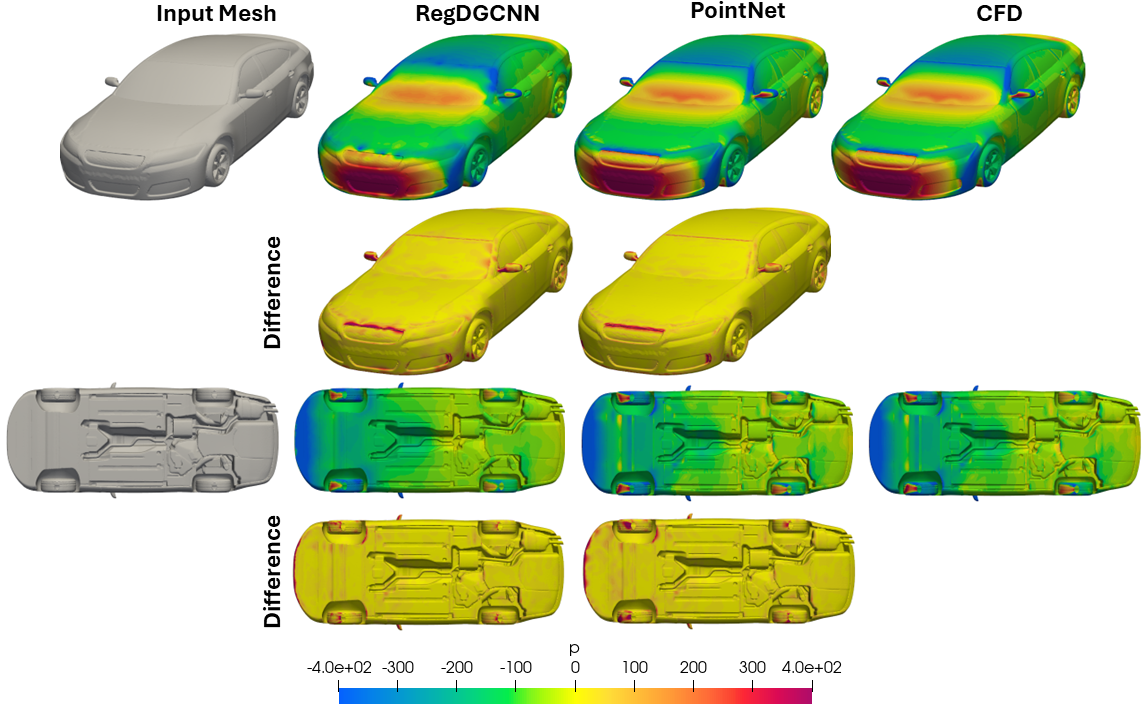}
\caption{Comparison of the input 3D mesh to RegDGCNN and PointNet for surface field predictions. The figure shows the input mesh (left), the prediction results from RegDGCNN (middle left), PointNet (middle right), CFD results as ground truth (right), and the difference between the model predictions and the ground truth from CFD simulations (bottom row).}
    \label{fig:PressurePredictions}
    \vspace{-10pt}
\end{figure}

In Figure \ref{fig:PressurePredictions}, we compare the input 3D mesh used by a modified PointNet~\cite{qi2017pointnet} and RegDGCNN~\cite{wang2019dynamic, elrefaie2024drivaernet} for predicting surface fields. PointNet is trained on 100k points, while RegDGCNN is limited to 10k points due to memory constraints. The figure shows the initial input mesh, the predicted results from both models, and the differences between these predictions and the ground truth from high-fidelity CFD simulations. This comparison illustrates the effectiveness of both models in capturing surface field variations and highlights the discrepancies between the predictions and actual CFD results, providing insights into each model's strengths and limitations.

For interpolation, we used Radial Basis Function (RBF) interpolation. The results summarized in Table~\ref{tab:L2_errors} present the L2 errors for the PointNet and RegDGCNN models in predicting surface fields. The L2 Error (Prediction vs Ground Truth) indicates how well the models' predictions align with the CFD data, while the L2 Error (Interpolated vs Full Mesh) evaluates the accuracy of the Radial Basis Function (RBF) interpolation against the full mesh data. For the PointNet model, the L2 error for predictions is 18.97\% with 100k nodes, and the error increases to 24.84\% with 350k nodes for the interpolated results. 
In comparison, the RegDGCNN model exhibits an L2 error of 24.66\% with 10k nodes for predictions and 26.44\% with 350k nodes for the interpolated mesh. These results suggest that the PointNet model demonstrates better prediction accuracy than the RegDGCNN model, particularly with a larger number of nodes, highlighting its effectiveness in capturing the geometric details necessary for accurate surface field predictions.

\begin{table}[h!]
    \centering
    \scriptsize
        \caption{L2 Errors for PointNet and RegDGCNN models in predicting surface fields.}
    \begin{tabular}{ccc}
        \hline
        Model & L2 Error (Prediction vs Ground Truth)& L2 Error (Interpolated vs Full Mesh) \\
                \hline
        PointNet & 18.97\% (100k nodes) & 24.84\% (350k nodes) \\
        RegDGCNN & 24.66\% (10k nodes) & 26.44\% (350k nodes) \\
        \hline
    \end{tabular}

    \label{tab:L2_errors}
\end{table}

\subsection{t-SNE Visualization of the Design Space}

\begin{figure}[h]
    \centering
    \includegraphics[width=\textwidth]{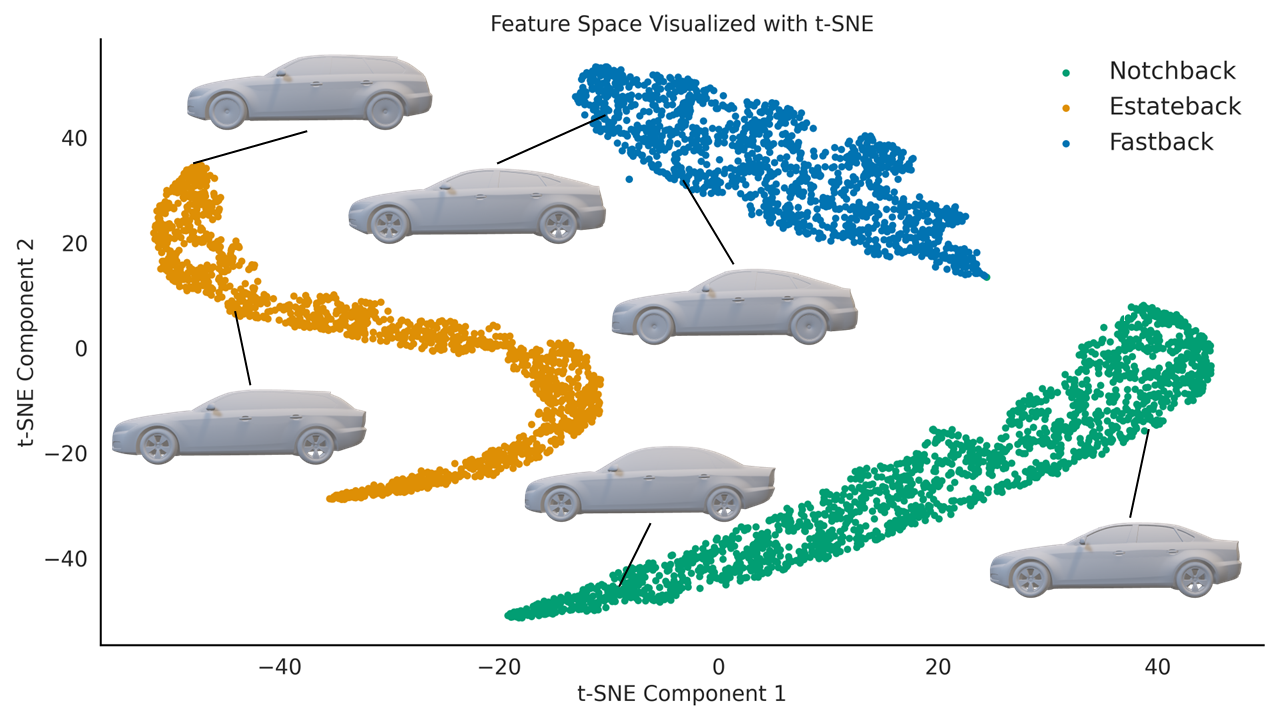}
    \caption{Feature space visualized with t-SNE for different car designs. The clusters represent three distinct car design categories: Notchback (green), Estateback (orange), and Fastback (blue). The t-SNE algorithm effectively separates the car designs into their respective categories, illustrating the ability of the feature representation to capture the underlying differences between the designs.}
    \label{fig:tsne_DrivAerNet}
\end{figure}

In this section, we train a PointNet~\cite{qi2017pointnet} model for 3D shape classification and utilize the learned features to visualize the design space. By reducing the dimensionality of these features to 2D using t-SNE (t-distributed stochastic neighbor embedding)~\cite{van2008visualizing}, we can create a plot that illustrates the similarity and variation between different car designs. This visualization aids designers in linking new designs to existing ones, potentially identifying the most similar designs and their performance using methods like K-Nearest Neighbors (KNN)~\cite{knn}. Using t-SNE to visualize the high-dimensional design space provides an efficient means to explore the vast design space. This method eliminates the need to compute complex metrics, such as the Chamfer distance, between a new design and all existing designs. By performing inference on a new design with a pretrained model, extracting the learned features, reducing their dimensionality, and visualizing their position in 2D space, we can easily assess how similar or different a new design is compared to existing ones. Figure~\ref{fig:tsne_DrivAerNet} demonstrates the t-SNE visualization, showing distinguished clusters for different car categories: Notchback, Estateback, and Fastback. Each point in the plot represents a unique design, illustrating the ability of t-SNE to capture and separate the underlying differences in car shapes effectively. In general, t-SNE is a powerful tool for reducing high-dimensional data to two or three dimensions while preserving the relative distances between data points, which helps in identifying patterns and similarities.

Furthermore, incorporating information about the aerodynamic performance of specific designs is crucial for analyzing new designs. Although Figure~\ref{fig:tsne_DrivAerNet} focuses on shape similarities, the methodology can be extended to include performance values, providing a more comprehensive understanding of different designs. This combined approach enables a more thorough analysis of design efficiency and performance, facilitating better-informed decisions in the design process.

\newpage
\small
\bibliographystyle{plainnat}
\bibliography{asmejour-sample} 
\normalsize

\clearpage 
\twocolumn
\definecolor{darkblue}{RGB}{46,25, 110}

\newcommand{\dssectionheader}[1]{%
   \noindent\framebox[\columnwidth]{%
      {\fontfamily{phv}\selectfont \textbf{\textcolor{darkblue}{#1}}}
   }
}

\newcommand{\dsquestion}[1]{%
    {\noindent \fontfamily{phv}\selectfont \textcolor{darkblue}{\textbf{#1}}}
}

\newcommand{\dsquestionex}[2]{%
    {\noindent \fontfamily{phv}\selectfont \textcolor{darkblue}{\textbf{#1} #2}}
}

\newcommand{\dsanswer}[1]{%
   {\noindent #1 \medskip}
}

\section*{Datasheets for Datasets}
In line with contemporary practices in the deep learning field, we employ datasheets to document our dataset's intent, structure, collection, preprocessing, labeling, usage, distribution, and maintenance, adhering to the established methodology referenced in~\cite{gebru2021datasheets}.
\newline
\begin{singlespace}

\dssectionheader{Motivation}

\dsquestionex{For what purpose was the dataset created?}{Was there a specific task in mind? Was there a specific gap that needed to be filled? Please provide a description.}

\dsanswer{The DrivAerNet++ dataset was developed to serve as a benchmark for training deep learning models.  This dataset, characterized by its extensive collection of 3D shapes, leverages $k$-$\omega$-SST RANS model to ensure high-fidelity turbulence modeling. Its significance lies in addressing the scarcity of publicly available, industry-grade datasets for training deep learning models in engineering design and CFD.}

\dsquestion{Who created this dataset (e.g., which team, research group) and on behalf of which entity (e.g., company, institution, organization)?}

\dsanswer{The dataset was created collaboratively by the Design Computation and Digital Engineering (DeCoDE) Lab at the Massachusetts Institute of Technology (MIT), the 3D AI Lab at the Technical University of Munich (TUM), and BETA CAE SYSTEMS USA, Inc.}

\dsquestionex{Who funded the creation of the dataset?}{If there is an associated grant, please provide the name of the grantor and the grant name and number.}

\dsanswer{The creation of this dataset was significantly supported by computational resources provided by MIT.}

\bigskip
\dssectionheader{Composition}

\dsquestionex{What do the instances that comprise the dataset represent?}{ Are there multiple types of instances? Please provide a description.}
\dsanswer{Each entry in the DrivAerNet++ dataset encompasses comprehensive 3D fields of velocity, pressure, wall-shear stresses, and aerodynamic coefficients, along with detailed 3D meshes of car bodies, front, and rear wheels. The fluid dynamics data is stored in \texttt{.vtk} files, while the geometric meshes are kept in \texttt{.stl} format, all optimized for high-speed data access. The dataset also includes parametric data saved in tabular format, point clouds, and part annotations as well as class labels for different car categories. To assist in efficient data utilization, a Python script is provided, enabling the conversion of these datasets into a format readily usable for machine learning applications, ensuring a smooth integration into various analysis pipelines.}

\dsquestion{How many instances are there in total (of each type, if appropriate)?}

\dsanswer{DrivAerNet++ contains a total of 8,000 3D full-domain CFD simulations.
}

\dsquestionex{Does the dataset contain all possible instances or is it a sample (not necessarily random) of instances from a larger set?}{ If the dataset is a sample, then what is the larger set? Is the sample representative of the larger set (e.g., geographic coverage)? If so, please describe how this representativeness was validated/verified. If it is not representative of the larger set, please describe why not (e.g., to cover a more diverse range of instances, because instances were withheld or unavailable).}

\dsanswer{Yes, the DrivAerNet++ dataset encompasses the complete 3D spatial domain as defined during the simulation process.}

\dsquestionex{What data does each instance consist of? “Raw” data (e.g., unprocessed text or images) or features?}{In either case, please provide a description.}

\dsanswer{Each instance primarily consists of velocity and pressure fields, turbulence metrics, boundary conditions, meshes, and log files. Additionally, it features derived data from functions such as \texttt{forceCoeffsIncompressible} for aerodynamic coefficients, \texttt{residuals} for equation residuals, and \texttt{wallShearStress}, along with visualization outputs like \texttt{streamlinesSphere} and \texttt{cutPlaneSurface} for flow and pressure visualization, offering a comprehensive overview of the fluid dynamics simulation.
}

\dsquestionex{Is there a label or target associated with each instance?}{If so, please provide a description.}

\dsanswer{Yes, all the aerodynamic coefficients ($C_d$, $C_l$, $C_{l,r}$, $C_{l,f}$, and $C_m$) are clearly labeled and ready for use in regression tasks. Furthermore, the flow-field data can be leveraged to generate additional labels for specific applications, such as the total pressure coefficient $C_{pt}$ and other quantities of interest. In addition, cars are labeled by design category and for each car we provide the labels of the components (part annotations).}

\dsquestionex{Are there recommended data splits (e.g., training, development/validation, testing)?}{If so, please provide a description of these splits, explaining the rationale behind them.}

\dsanswer{Yes, DrivAerNet++ is designed as a benchmark dataset for ML applications, complete with a split into training, validation, and test sets to ensure fair comparisons among various machine and deep learning models. Details on the data splits can be found here: \url{https://github.com/Mohamedelrefaie/DrivAerNet}.
}

\dsquestionex{Are there any errors, sources of noise, or redundancies in the dataset?}{If so, please provide a description.}

\dsanswer{Yes. Reynolds-Averaged Navier-Stokes (RANS) turbulence models have been utilized. Despite the employment of these models, inherent limitations and approximations introduce potential inaccuracies. RANS models, particularly effective for a broad spectrum of flow conditions, may not capture the full range of turbulent scales and complex flow phenomena with absolute precision. Drawbacks stem from the models' reliance on empirical data and simplifying assumptions, which can lead to discrepancies in predicting highly unsteady or anisotropic turbulent flows. These limitations necessitate cautious interpretation of the data, especially in scenarios where turbulence plays a critical role. See~\cite{islam2009application} and~\cite{ashton2016assessment} for more details.}

\dsquestionex{Is the dataset self-contained, or does it link to or otherwise rely on external resources (e.g., websites, tweets, other datasets)?}{If it links to or relies on external resources, a) are there guarantees that they will exist, and remain constant, over time; b) are there official archival versions of the complete dataset (i.e., including the external resources as they existed at the time the dataset was created); c) are there any restrictions (e.g., licenses, fees) associated with any of the external resources that might apply to a future user? Please provide descriptions of all external resources and any restrictions associated with them, as well as links or other access points, as appropriate.}

\dsanswer{Our dataset is self-contained.}

\dsquestionex{Does the dataset contain data that might be considered confidential (e.g., data that is protected by legal privilege or by doctor-patient confidentiality, data that includes the content of individuals non-public communications)?}{If so, please provide a description.}

\dsanswer{No.
}

\bigskip
\dssectionheader{Collection Process}

\dsquestionex{What mechanisms or procedures were used to collect the data (e.g., hardware apparatus or sensor, manual human curation, software program, software API)?}{How were these mechanisms or procedures validated?}

\dsanswer{For the DrivAerNet++ dataset, we predominantly utilized open-source tools to enable researchers to replicate the methodology and approach. The simulations were conducted using OpenFOAM®~\cite{OpenFOAMv11}, while SnappyHexMesh was employed for mesh generation. Geometry processing was handled by Blender~\cite{blender} and the commercial software ANSA®. Python scripts facilitated the automation of the processes, and ParaView was used for visualization and data analysis. Although we used the commercial software ANSA® for creating the parametric model and conducting the design of experiments, we are releasing the scripts used in ANSA® along with the parametric models. This ensures that results can be replicated using any CAD modeling software, facilitating reproducibility.}

\dsquestion{Who was involved in the data collection process (e.g., students, crowdworkers, contractors) and how were they compensated (e.g., how much were crowdworkers paid)?}

\dsanswer{The DrivAerNet++ dataset was developed and refined solely by the authors of this work without the involvement of crowdworkers or contractors, and no extra compensation was provided beyond the usual salary and stipend.}

\dsquestionex{Over what timeframe was the data collected? Does this timeframe match the creation timeframe of the data associated with the instances (e.g., recent crawl of old news articles)?}{If not, please describe the timeframe in which the data associated with the instances was created.}

\dsanswer{The data was collected in Q3, and Q4 of 2023 as well as Q1, and Q2 of 2024.
}

\bigskip
\dssectionheader{Preprocessing/cleaning/labeling}

\dsquestionex{Was any preprocessing/cleaning/labeling of the data done (e.g., discretization or bucketing, tokenization, part-of-speech tagging, SIFT feature extraction, removal of instances, processing of missing values)?}{If so, please provide a description. If not, you may skip the remainder of the questions in this section.}

\dsanswer{Yes, preprocessing was essential due to the wide range of morphing and geometry modifications applied. Some of the resulting geometries were unsuitable for CFD simulations due to issues such as intersecting faces and non-manifold edges. Additionally, some modifications led to unrealistic car shapes that deviated significantly from practical automotive designs. To address these issues, we implemented a validation process to identify and exclude such problematic geometries, ensuring that only feasible designs were included for simulations.}

\dsquestionex{Was the “raw” data saved in addition to the preprocessed/cleaned/labeled data (e.g., to support unanticipated future uses)?}{If so, please provide a link or other access point to the “raw” data.}

\dsanswer{Yes, the raw data from the OpenFOAM® simulations were saved. However, due to the considerable size of the dataset, it will be shared upon request and subject to approval from MIT. This ensures accessibility while maintaining proper management and oversight.}

\dsquestionex{Is the software used to preprocess/clean/label the instances available?}{If so, please provide a link or other access point.}

\dsanswer{Yes, we employed open-source tools including Python, Blender, and ParaView for preprocessing, cleaning, and labeling the dataset instances.}

\bigskip
\dssectionheader{Uses}

\dsquestionex{Has the dataset been used for any tasks already?}{If so, please provide a description.}

\dsanswer{Yes, the dataset has been utilized in our study focusing on the surrogate modeling of the aerodynamic drag and surface fields prediction, with findings documented in the paper.}

\dsquestion{What (other) tasks could the dataset be used for?}

\dsanswer{The DrivAerNet++ dataset could be utilized for a variety of tasks including data-driven aerodynamic design optimization, turbulence modeling, reduced-order modeling, generative AI, dimensionality reduction, and aerodynamic design analysis.
}

\bigskip
\dssectionheader{Distribution}

\dsquestionex{Will the dataset be distributed to third parties outside of the entity (e.g., company, institution, organization) on behalf of which the dataset was created?}{If so, please provide a description.}

\dsanswer{Yes, the dataset is publicly available at \url{https://github.com/Mohamedelrefaie/DrivAerNet}
}

\dsquestionex{How will the dataset will be distributed (e.g., tarball on website, API, GitHub)}{Does the dataset have a digital object identifier (DOI)?}

\dsanswer{A Digital Object Identifier (DOI) will be assigned and shared alongside the dataset release on GitHub.}

\dsquestion{When will the dataset be distributed?}

\dsanswer{The complete dataset will be released after our study has been peer-reviewed and accepted for publication. Meanwhile, a smaller portion of the dataset will be released for initial public use and for replicating the results in the main paper.}

\dsquestionex{Will the dataset be distributed under a copyright or other intellectual property (IP) license, and/or under applicable terms of use (ToU)?}{If so, please describe this license and/or ToU, and provide a link or other access point to, or otherwise reproduce, any relevant licensing terms or ToU, as well as any fees associated with these restrictions.}

\dsanswer{The dataset is provided under the Creative Commons Attribution-NonCommercial (CC BY-NC) license, allowing use and modification for non-commercial purposes provided appropriate credit is given. No fees are associated with its use. Full terms can be found at \url{https://creativecommons.org/licenses/by-nc/4.0/deed.en}, ensuring accessibility while acknowledging creators' contributions.}

\bigskip
\dssectionheader{Maintenance}

\dsquestion{Who will be supporting/hosting/maintaining the dataset?}

\dsanswer{The dataset is maintained by the DeCoDE Lab at MIT. The DrivAerNet team is committed to maintaining the dataset by addressing issues reported on our GitHub issues page. The dataset will be hosted on the Harvard Dataverse Repository for long-term availability, and our data management approach aligns with the FAIR principles~\cite{wilkinson2016fair} for responsible scientific data handling.
}

\dsquestion{How can the owner/curator/manager of the dataset be contacted (e.g., email address)?}

\dsanswer{The DrivAerNet++ team can be reached at \url{https://decode.mit.edu/projects/DrivAerNet/} . Users are encouraged to submit inquiries to the GitHub page. For further inquiries and collaborations, Mohamed Elrefaie can be contacted directly.
}

\dsquestionex{Is there an erratum?}{If so, please provide a link or other access point.}

\dsanswer{Yes. A comprehensive issue-tracking system is available at 
\url{https://github.com/Mohamedelrefaie/DrivAerNet/issues}.
}

\dsquestionex{Will the dataset be updated (e.g., to correct labeling errors, add new instances, delete instances)?}{If so, please describe how often, by whom, and how updates will be communicated to users (e.g., mailing list, GitHub)?}

\dsanswer{Yes, the team plans to issue minor updates for the dataset to address any discovered errors. These updates will be announced on our GitHub page.
}

\dsquestionex{If others want to extend/augment/build on/contribute to the dataset, is there a mechanism for them to do so?}{If so, please provide a description. Will these contributions be validated/verified? If so, please describe how. If not, why not? Is there a process for communicating/distributing these contributions to other users? If so, please provide a description.}

\dsanswer{Yes, we welcome contributions through creating pull requests at \url{https://github.com/Mohamedelrefaie/DrivAerNet/pulls}. All submissions will be reviewed for accuracy, and reliability before being added to the dataset. Changes from new contributions will be recorded in our version history.}

\end{singlespace}


\end{document}